\documentclass[]{bytedance_seed}

\usepackage{hyperref}
\usepackage{cleveref}
\usepackage{verbatim}

\usepackage{wrapfig}  
\usepackage{graphicx}
\usepackage{subcaption}
\usepackage{listings}
\usepackage{algorithm}
\usepackage{amsmath, amssymb}   
\usepackage{booktabs}           
\usepackage{multirow}           
\usepackage{makecell}           
\usepackage{fontawesome5}

\usepackage[table]{xcolor}

\usepackage{subcaption} %

\usepackage[toc,page,header]{appendix}

\usepackage{minitoc}

\title{BitDance: Scaling Autoregressive Generative Models with Binary Tokens}

\author{%
\parbox{\textwidth}{\centering
Yuang Ai$^{*1,2,4}$, Jiaming Han$^{*1,2}$, Shaobin Zhuang$^{*1,3}$, Weijia Mao$^{1,5}$, Xuefeng Hu$^{1}$\\[1mm]
Ziyan Yang$^{1}$, Zhenheng Yang$^{1}$, Yali Wang$^{6}$, Huaibo Huang$^{4\dagger}$, Xiangyu Yue$^{2\dagger}$\\[1mm] 
Hao Chen$^{*1\dagger\ddagger}$
}}

\affiliation{%
\parbox{\textwidth}{\centering\small
$^1$ByteDance,
$^2$MMLab, The Chinese University of Hong Kong,
$^3$Shanghai Jiao Tong University\\[1mm]
$^4$Institute of Automation, Chinese Academy of Sciences,
$^5$National University of Singapore\\[1mm]
$^6$Shenzhen Institutes of Advanced Technology, Chinese Academy of Sciences
\vspace{-1mm}

}}

\contribution[*]{Equal contribution}
\contribution[\dagger]{Corresponding Author}
\contribution[\ddagger]{Project lead}

\abstract{
We present BitDance, a scalable autoregressive (AR) image generator that predicts binary visual tokens instead of codebook indices.
With high-entropy binary latents, BitDance lets each token represent up to $2^{256}$ states, yielding a compact yet highly expressive discrete representation.
Sampling from such a huge token space is difficult with standard classification. To resolve this, BitDance uses a binary diffusion head: instead of predicting an index with softmax, it employs continuous-space diffusion to generate the binary tokens.
Furthermore, we propose next-patch diffusion, a new decoding method that predicts multiple tokens in parallel with high accuracy, greatly speeding up inference.
On ImageNet 256×256, BitDance achieves an FID of 1.24, the best among AR models. With next-patch diffusion, BitDance beats state-of-the-art parallel AR models that use 1.4B parameters, while using 5.4× fewer parameters (260M) and achieving 8.7× speedup.
For text-to-image generation, BitDance trains on large-scale multimodal tokens and generates high-resolution, photorealistic images efficiently, showing strong performance and favorable scaling. When generating 1024×1024 images, BitDance achieves a speedup of over 30× compared to prior AR models. We release code and models to facilitate further research on AR foundation models. 
}

\checkdata[\faGithub\ Code]{\url{https://github.com/shallowdream204/BitDance}}

\checkdata[\faGlobe\ Project Page]{\url{https://bitdance.csuhan.com}}

\begin{document}
\maketitle

\begin{figure}[H] 
\centering
\vspace{-10mm}
    \includegraphics[width=0.9\textwidth]{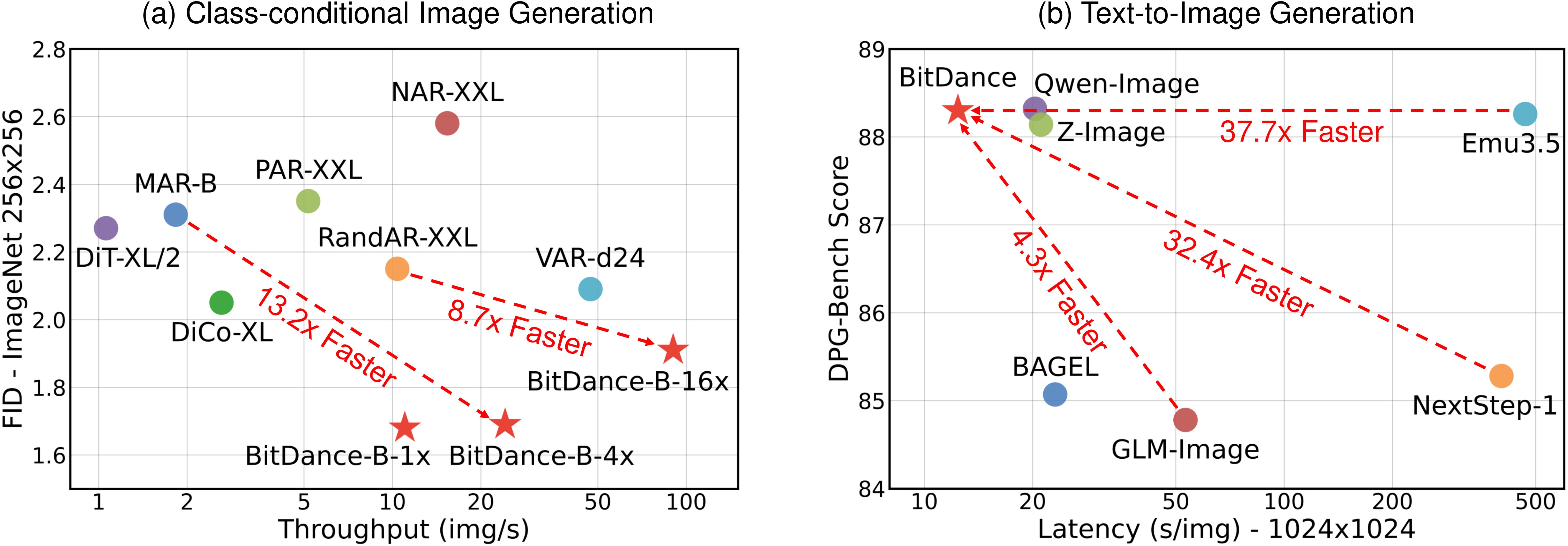}
    \vspace{-2mm}
    \caption{\textbf{
    Performance vs. efficiency compared with SOTA diffusion models and autoregressive models.
    }} 
    \label{fig:speed}
    \vspace{-0.4cm}
\end{figure}

\begin{figure}[H] 
\centering
    \includegraphics[width=1.0\textwidth]{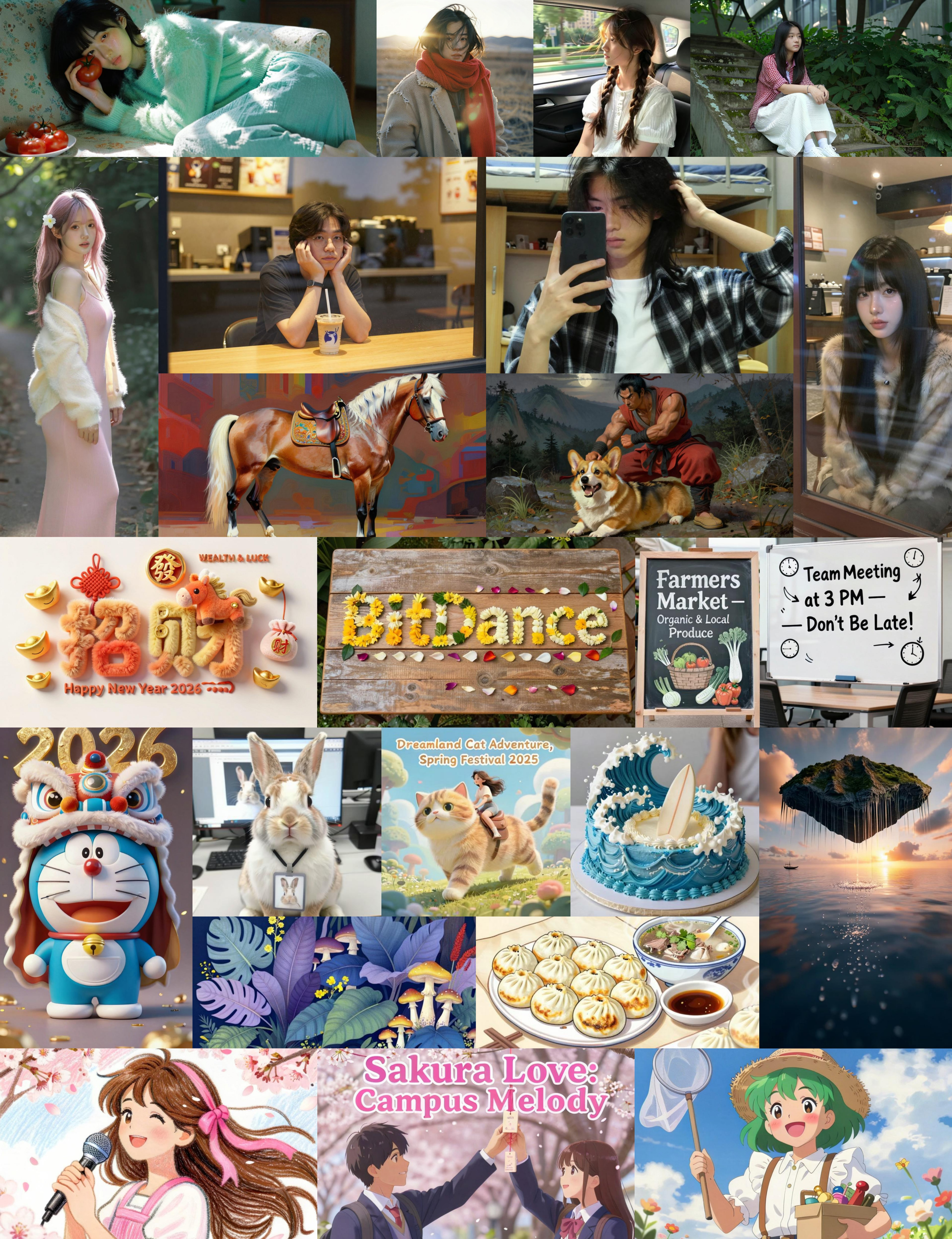}
    \caption{\textbf{High-resolution samples generated by the 14B BitDance model}, showcasing its capabilities in prompt
adherence, spatial reasoning, and text rendering across various aspect ratios and artistic styles.} 
    \label{fig:teaser}%
\end{figure}

\section{Introduction}

Large language models (LLMs) have shown that autoregressive pre-training via next-token prediction can distill large-scale textual knowledge into a single model with strong generalization~\cite{achiam2023gpt,comanici2025gemini,guo2025deepseek}. This success has motivated extending AR modeling to visual generation~\cite{sun2024autoregressive,tian2024visual,li2024autoregressive}, yielding a unified paradigm with language modeling and enabling progress on multimodal applications~\cite{fan2024fluid,team2025nextstep,glmimage}.

However, applying the AR paradigm to vision still faces persistent challenges. A primary issue is token design: visual tokens must be expressive enough to capture rich image content while remaining sufficiently regularized to limit error accumulation over long sequences.
Existing discrete AR models typically leverage Vector Quantization (VQ)~\cite{van2017neural,esser2021taming} for tokenization, but often suffer from degraded reconstruction due to difficulties in scaling visual vocabularies~\cite{sun2024autoregressive,tian2024visual,wang2024emu3}. In contrast,  continuous AR models use VAEs to achieve superior reconstruction, the unconstrained nature of their latent spaces often leads to severe error accumulation~\cite{xar,ke2025hyperspherical,team2025nextstep}. Furthermore, the sequential nature of token-by-token generation imposes a significant bottleneck on inference efficiency, particularly as image resolutions scale upward.

In this paper, we present BitDance, a simple yet scalable autoregressive framework that achieves state-of-the-art performance in image generation. BitDance is built upon three key components: \textbf{(i)} a large-vocabulary binary tokenizer, \textbf{(ii)} a binary diffusion head for sampling in extremely large discrete spaces, and \textbf{(iii)} a next-patch diffusion paradigm that enables efficient multi-token prediction. 

Inspired by recent advances in binary quantization~\cite{yulanguage,zhao2024image}, we scale the entropy of binary representations, expanding the vocabulary size up to $2^{256}$. This scale is orders of magnitude larger than that of existing discrete tokenizers \cite{sun2024autoregressive,han2025infinity,wang2025bridging}, enabling our discrete representation to surpass continuous VAEs in reconstruction fidelity (see Tab.~\ref{tab:rec_in1k}). By encoding images into a compact yet expressive binary latent space, BitDance preserves fine-grained visual details while introducing beneficial discreteness that helps regularize long-sequence generation and mitigate error accumulation.

\begin{figure}[t] 
\centering
    \includegraphics[width=0.95\textwidth]{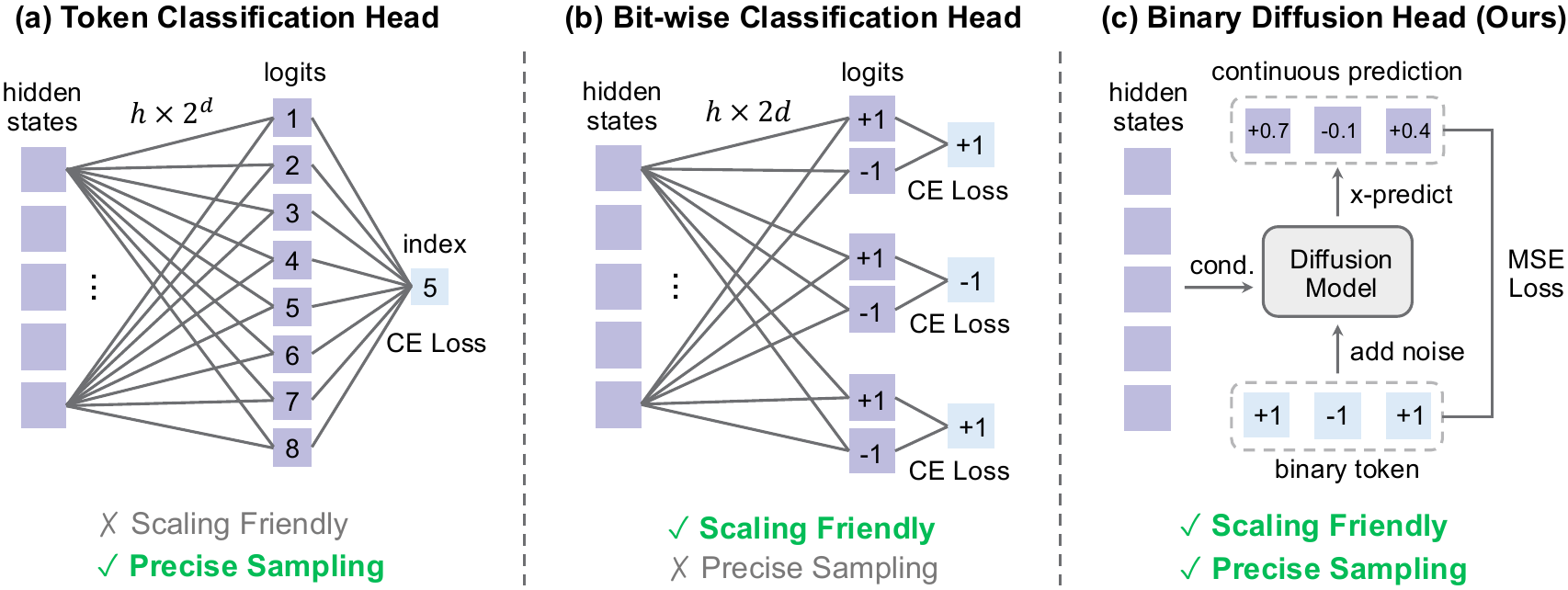}
    \caption{\textbf{Comparison of binary token sampling paradigms.} Scaling up binary token entropy yields reconstruction performance on par with continuous VAEs, but it simultaneously creates a bottleneck during sampling. For a $d$-channel binary token: (a) Directly modeling $p(b_1,b_2,\dots,b_d)$ requires $h\times2^d$ parameters, which suffers from an exponential explosion as $d$ scales. (b) Bit-wise classification~\cite{han2025infinity} reduces the parameter count to $h\times2d$ by assuming bit independence, i.e., ${\textstyle \prod_{i=1}^{d}} p(b_i)$, but this restrictive assumption compromises sampling fidelity. (c) We embed binary tokens as vertices of a $d$-dimensional hypercube in continuous space. By modeling the joint distribution of all bits via a diffusion objective, we achieve controllable parameters as $d$ scales up and high-fidelity sampling.} 
    \label{fig:head}
\end{figure}

However, such an expansive vocabulary poses severe challenges for conventional sampling. As shown in Fig.~\ref{fig:head}, index-based classification heads suffer from an inherent trade-off between parameter efficiency and sampling accuracy. To address this issue, we propose a binary diffusion head. Rather than mapping binary tokens to discrete indices, we embed binary tokens as vertices of a hypercube in continuous space and model their distribution using a diffusion objective~\cite{ho2020denoising}. By jointly modeling all binary channels, the proposed head enables accurate sampling while maintaining a controllable parameter footprint.

To further accelerate vision token prediction, we propose a next-patch diffusion approach for visual autoregressive modeling. Considering the spatial dependency in images, we posit that tokens within a local patch are highly correlated, making them easier to predict jointly. As shown in Fig.~\ref{fig:parallel}, unlike prior parallel AR generation methods~\cite{wang2025parallelized,pang2025randar,li2025autoregressive} that use factorized sampling—sampling each token independently via a classification head—BitDance uses a binary diffusion head to explicitly model the joint distribution of tokens generated in parallel. This design enables high-fidelity parallel token prediction and substantially improves efficiency.

BitDance demonstrates superior performance in both class-conditional and text-to-image generation. On ImageNet 256$\times$256, the 1B BitDance model achieves an FID of 1.24, outperforming previous AR models. As shown in Fig.~\ref{fig:speed}, BitDance outperforms 1.4B state-of-the-art parallel AR generative model using only 260M parameters, while achieving an 8.7$\times$ speedup. For text-to-image generation, we scale BitDance to 14B parameters and train on large-scale multi-modal tokens. After pre-training, continued training, and supervised fine-tuning, BitDance exhibits strong text-to-image synthesis capabilities, generating high-resolution images with fine-grained details. Specifically, BitDance achieves scores of 0.86 on GenEval~\cite{ghosh2023geneval}, 88.28 on DPG-Bench~\cite{hu2024ella}, 0.532 on OneIG-EN~\cite{chang2025oneig}, and 0.512 on OneIG-ZH~\cite{chang2025oneig}, setting a new state-of-the-art among existing autoregressive models. As shown in Fig.~\ref{fig:speed}, our model achieves a speedup of over 30× when generating 1024×1024 images, compared to standard next-token prediction AR models (NextStep-1~\cite{team2025nextstep} and Emu3.5~\cite{cui2025emu35}).

Our main contributions can be summarized as follows:
\begin{itemize}
    \item We present BitDance, a simple and scalable autoregressive model. BitDance demonstrates the viability of scaling token entropy for high-fidelity visual generation, offering new insights into the design space of visual autoregressive modeling.
    \item We propose a binary diffusion head to resolve the sampling bottlenecks inherent in expansive visual vocabularies. By seamlessly extending it to multi-token sampling, BitDance facilitates precise and efficient parallel prediction through a next-patch diffusion paradigm.
    \item Extensive experiments on class-conditional and text-to-image generation show that BitDance offers exceptional scaling, achieving superior generative quality with high inference speed.
\end{itemize}

\section{Related Work}

\subsection{Visual Tokenizers}
To mitigate the prohibitive costs of pixel-space training~\cite{chen2020generative,ho2020denoising}, Variational Autoencoders (VAEs)~\cite{kingma2013auto} are widely used to project visual content into continuous latent spaces. Due to their high-fidelity reconstruction and smooth gradient flow, VAEs have become the standard for leading diffusion models~\cite{rombach2022high,dit,flux,xie2025sana}. Conversely, discrete tokenizers using Vector Quantization (VQ)~\cite{van2017neural,esser2021taming} often struggle with quantization errors and instability of codebook utilization. Recent works have increasingly investigated binary quantization as a solution. MAGVIT-v2~\cite{yulanguage} introduces Lookup-Free Quantization (LFQ) to scale vocabularies to $2^{18}$, but its entropy loss incurs linear memory costs that hinder further scaling. Subsequent works like BSQ~\cite{zhao2024image} and WeTok~\cite{zhuang2025wetok} attempt to resolve this via independence assumptions or grouping strategies, respectively. Building on these advances, we aim to further explore scaling up the vocabulary size to achieve higher token entropy.

\subsection{Autoregressive Viusal Generation}
Standard autoregressive (AR) visual generation typically quantizes images into discrete tokens and models their distribution via next-token prediction in a raster-scan order~\cite{sun2024autoregressive,wang2024emu3,cui2025emu35}. Recently, several paradigm shifts have emerged, exploring continuous token spaces~\cite{li2024autoregressive,ren2025flowar}, randomized ordering~\cite{yu2025randomized,pang2025randar,li2025autoregressive}, and alternative modeling primitives~\cite{tian2024visual,xar}. For instance, MAR~\cite{li2024autoregressive} introduces a token-level diffusion head to facilitate continuous token sampling, while NextStep-1~\cite{team2025nextstep} scales up the continuous AR framework for high-fidelity text-to-image synthesis. However, continuous tokens often lack sufficient regularization, leading to severe error accumulation and representation drift~\cite{xar,sun2024multimodal} during long-sequence generation—factors that significantly degrade the quality of high-resolution images. SphereAR~\cite{ke2025hyperspherical} employs hyperspherical constraints to regularize the latent features of VAEs. In this work, we focus on discrete binary representations to explore the potential of this more radical constraint in autoregressive visual generation.

\subsection{Parallel Prediction in AR Models}
Accelerating AR generation has become a pivotal research direction in visual generation. Mask-GIT~\cite{chang2022maskgit} and MAR~\cite{li2024autoregressive} adopt MAE-style masking strategies for modeling. VAR~\cite{tian2024visual} utilizes next-scale prediction to predict tokens within a unified scale in parallel. PAR~\cite{wang2025parallelized} employs a grouping strategy to generate weakly dependent tokens. RandAR~\cite{pang2025randar} and ARPG~\cite{li2025autoregressive} leverage random-order modeling to enable the prediction of tokens at arbitrary positions. While showing promise, these methods often struggle to model the joint distribution of tokens predicted in parallel, lacking sufficient multi-token constraints during the final sampling stage (see Fig.~\ref{fig:parallel}). Our BitDance seamlessly extends the binary diffusion head to model the joint distribution of multi-token, achieving efficient and reliable parallel prediction through next-patch diffusion.

\section{BitDance}

\subsection{Binary Visual Tokenizer}

For discrete visual tokenizers, scaling up the vocabulary size to increase token entropy is critical for enhancing both reconstruction fidelity and downstream generation quality. However, traditional Vector Quantization (VQ) often encounters codebook collapse as the vocabulary expands. To ensure efficient utilization of the codebook space, we adopt binary quantization via Lookup-Free Quantization (LFQ)~\cite{yulanguage}. Given an encoded latent token $x\in \mathbb{R}^{d}$, LFQ employs an implicit, learning-free codebook $\mathcal{C}_{LFQ}=\{-1,1\}^d$. The binary quantization process is performed independently across each channel of $x$:
\begin{equation}
    x_{q} = \mathrm{sign}(x).
    \label{eq:lfq}
\end{equation}
To prevent codebook collapse and maximize information capacity, an entropy loss~\cite{Jansen2019CoincidenceCA} is typically employed:
\begin{equation}
    \mathcal{L}_{entropy} = \mathbb{E}[H(q(x)]) -H[\mathbb{E}(q(x))],
\end{equation}
where $H(\cdot)$ denotes the entropy calculation. In standard LFQ, computing the distribution $q(x)$ requires calculating similarities between $x$ and the entire codebook space. As the vocabulary size grows exponentially ($2^d$), this calculation becomes computationally prohibitive due to its significant memory footprint.

To address this bottleneck,  we adopt a group-wise LFQ strategy~\cite{zhuang2025wetok}, which partitions $d$ channels into $g$ distinct groups for entropy calculation. This strategy strikes a balance between computational efficiency and optimization accuracy, enabling the training of models with an exceptionally large vocabulary. Specifically, we scale the codebook size up to $\mathbf{2^{256}}$. The vast representation space allows our discrete tokenizer to achieve reconstruction fidelity comparable to continuous VAEs (see Tab.~\ref{tab:rec_in1k}). 

\begin{figure*}[t] 
\centering
    \includegraphics[width=1.0\textwidth]{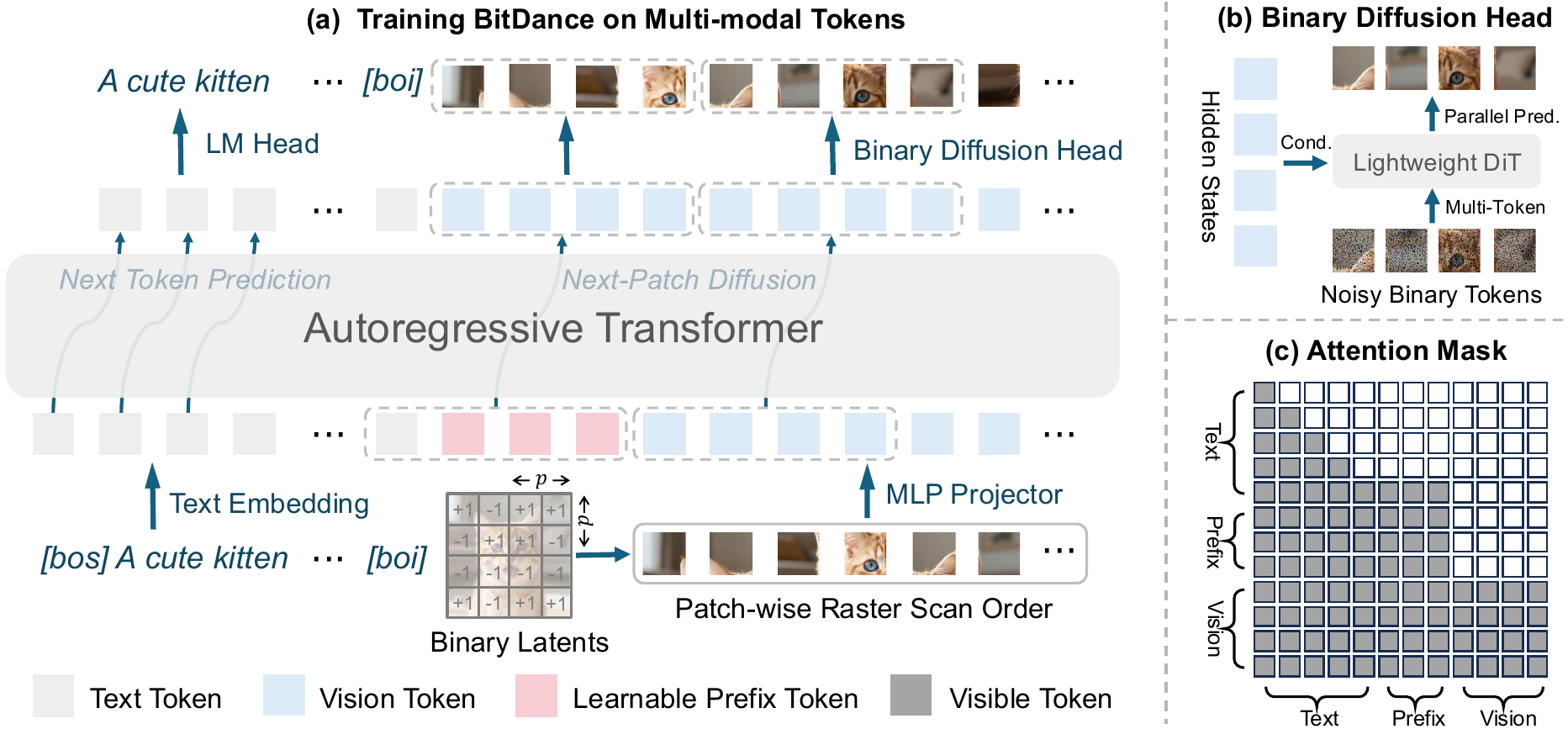}
    \caption{\textbf{Architecture of BitDance,} an autoregressive model trained on multi-modal tokens. An input image is first encoded into binary latents and then flattened into a 1D sequence following a patch-wise raster scan order with patch size $p\times p$. Vision tokens are modeled using our proposed next-patch diffusion, utilizing a binary diffusion head to achieve efficient and precise parallel prediction.} 
    \label{fig:arch}
\end{figure*}

\subsection{Binary Diffusion Head}
\label{sec:head}

While a large vocabulary increases token entropy and reconstruction fidelity, it introduces challenges for sampling. For a binary token with $d$ channels, the number of possible discrete indices grows exponentially to $2^d$. As illustrated in Fig.~\ref{fig:head}, conventional index-based classifiers face an inherent trade-off between sampling precision and parameter efficiency.

\textbf{The Sampling Bottleneck.} To model the joint probability $p(b_1,b_2,\dots,b_d)$ of a $d$-bit token, a standard classification head must characterize a categorical distribution across all $2^d$ possible indices. This approach incurs a prohibitive parameter overhead of $h\times2^d$, where $h$ denotes the hidden dimension. As $d$ scales, this becomes computationally intractable; for instance, with $h=1024$ and $d=32$, the parameter count reaches approximately 4.4 trillion, far exceeding the capacity of modern hardware.

To address this, \cite{han2025infinity} adopts a bit-wise independence assumption, decomposing the joint distribution into $d$ separate binary classifications. By modeling the distribution as a product of independent probabilities, ${\textstyle \prod_{i=1}^{d}} p(b_i)$, the approach effectively reduces the parameter overhead to $h\times 2d$. However, this assumption is overly restrictive, as it fails to capture the intricate inter-bit correlations in binary tokens, leading to degraded sampling precision and inferior generative quality (see Tab.~\ref{tab:abla_head}).

\textbf{Diffusion-based Binary Prediction.} We propose a novel binary diffusion head to address these sampling difficulties. Unlike conventional applications of diffusion models that primarily focus on modeling continuous data distributions, we are the first to leverage such models for discrete binary tokens. Rather than mapping these tokens to discrete indices, we represent them as $d$-dimensional vectors within a continuous space. Specifically, let $x\in\mathbb{R}^d$ denote the ground-truth binary token to be predicted and $z\in\mathbb{R}^h$ represent the condition (i.e., the hidden state from the AR transformer). To model the conditional probability distribution $p(x|z)$, we adopt the Rectified Flow~\cite{liuflow} formulation and optimize the head using $x$-prediction with a velocity-matching loss~\cite{li2025back}:
\begin{equation} 
\label{eq:diff_loss}
\mathcal{L}(z,x) = \mathbb{E}_{t, x, \epsilon}  \left\| v_{\theta}(x_t, t, z) - v_t \right\|^2, 
\end{equation}
where $x_t=tx+(1-t)\epsilon$ is the noisy token at time $t$ interpolated with Gaussian noise $\epsilon\sim\mathcal{N}(0,\mathbf{I})$, and $v_t=x-\epsilon$ is the target velocity. The velocity $v_\theta$ is parameterized by a $x$-prediction network $f_\theta$ such that $v_{\theta}(x_t,t,z) = (f_\theta(x_t,t,z)-x_t) / (1-t)$.

During inference, we initialize $x_0\sim\mathcal{N}(0,\mathbf{I})$ and integrate the learned velocity field using an Euler solver with $N$ uniform steps $\Delta_t=1/N$:
\begin{equation}
    x_{t+\Delta_t}=x_t+v_\theta(x_t,t,z)\Delta_t.
\end{equation}
After $N$ steps, we apply a \textbf{hard binarization constraint}: $x_1=\mathrm{sign}(x_1)$. This step effectively projects the continuous prediction back onto the binary hypercube, exploiting the structural priors of the token space. This mechanism circumvents the compounded error accumulation typically found in AR models that operate in unconstrained continuous spaces~\cite{ke2025hyperspherical, team2025nextstep}.

\textbf{Why Binary Diffusion Head is Effective.} Geometrically, binary tokens in continuous space form a finite set of vertices on a hypercube, characterized by uniform magnitudes but distinct orientations. Unlike standard VAE latent spaces which are often unconstrained, the binary target space provides a strong structural regularization. This restriction to a finite, well-defined set of target vectors significantly reduces the optimization complexity for the diffusion head, leading to faster convergence and superior sampling stability (see Tab.~\ref{tab:abla_vae}). Fig.~\ref{fig:head_distribution} illustrates the output distribution of the binary diffusion head. It reveals that the model effectively learns the discrete nature of binary tokens without any manual constraints.

\subsection{Next-Patch Diffusion}
Standard autoregressive (AR) generation formulates image synthesis as a sequence of next-token predictions. Formally, given a flattened 1D sequence of $N$ tokens $x=[x_1,x_2,\dots,x_N]$, the generative process is decomposed into a chain of conditional probabilities:
\begin{equation}
    p(x) = \prod_{n=1}^{N}  p(x_n|x_1,x_2,\dots,x_{n-1}).
    \label{eq:ar}
\end{equation}
While effective, this token-by-token paradigm imposes a significant inference bottleneck, particularly as image resolutions and sequence lengths increase.

\textbf{From Next-Token to Next-Patch.} To alleviate this, we exploit the inherent spatial dependency in visual data. We posit that tokens within a local patch exhibit strong statistical dependencies and can thus be predicted in parallel without compromising generative quality. In our next-patch prediction scheme, the sequence $x$ is partitioned into $M$ disjoint groups (patches), $x=[X_1,X_2,\dots,X_M]$, where each group $X_m=\{x_m^1,x_m^2,\dots,x_m^{p^2}\}$ contains $p\times p$ tokens.
The parallel AR generation can be expressed as:
\begin{equation} p(x) = \prod_{m=1}^{M} p(X_m \mid X_1, \dots, X_{m-1}). \label{eq:parallel_ar} \end{equation}
As illustrated in Fig.~\ref{fig:arch}, we adhere to a patch-wise raster-scan order. To implement this within the AR Transformer, we employ a block-wise causal attention mask. Unlike standard causal masks that enforce a strict 1D dependency, our block-wise configuration allows tokens within the same patch to be mutually visible. This enables the model to explicitly capture intra-patch spatial interactions while maintaining the autoregressive dependency across patches. Besides, to facilitate the simultaneous prediction of all tokens in the initial patch, we introduce $p^2-1$ learnable prefix tokens as placeholders before vision tokens.

\begin{figure}[t]
\centering
\hspace{-1.5em}
\begin{minipage}{0.6\linewidth}{
\includegraphics[width=0.95\textwidth]{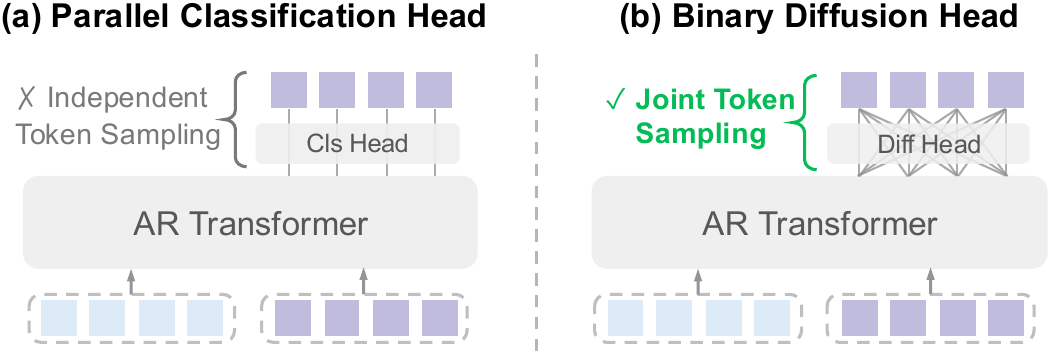}
}\end{minipage}
\begin{minipage}{0.4\linewidth}{
\caption{
\textbf{Comparison of different sampling heads for parallel prediction in autoregressive models.} (a) The standard classification head is limited to independent token sampling, which violates the inherent dependencies required for parallel prediction. (b) Our proposed binary diffusion head models the joint distribution of tokens generated simultaneously, enabling coherent sampling.
}
\label{fig:parallel}
}\end{minipage}
\end{figure}

\textbf{Precise Multi-token Sampling.} While recent parallel AR frameworks~\cite{wang2025parallelized, pang2025randar, li2025autoregressive} have introduced various modeling trajectories and specialized attention mechanisms, we contend that their efficacy is fundamentally constrained by a \textbf{training-inference discrepancy in the parallel AR objective}. Specifically, these methods still adhere to the token-wise training objectives defined in Eq.~(\ref{eq:ar}). As shown in Fig.~\ref{fig:parallel}, this creates a critical mismatch during inference: while the model is tasked with generating a group of tokens simultaneously, the tokens within that group are \textbf{sampled independently} through a standard classification head. Such an implicit independence assumption directly violates the \textbf{joint distribution modeling} necessitated by Eq.~(\ref{eq:parallel_ar}), inevitably compromising structural coherence and introducing artifacts within the generated groups.

To bridge this gap, we extend the binary diffusion head introduced in Sec.~\ref{sec:head} to support joint multi-token prediction. By leveraging the flexibility of the diffusion formulation, we adapt the training objective in Eq.~(\ref{eq:diff_loss}) to multi-token. Let $X\in\mathbb{R}^{p^2\times d}$ represent the patch of ground-truth tokens and $Z\in\mathbb{R}^{p^2\times h}$ denote the corresponding hidden states. The optimization objective becomes:

\begin{equation} \label{eq:diff_loss_patch} \mathcal{L}_{parallel} = \mathbb{E}_{t, X, \epsilon} \left|| v_{\theta}(X_t, t, Z) - v_t \right||^2. \end{equation}

To effectively model the $p^2$ tokens within the head, we design the architecture of prediction network $f_\theta$ as a lightweight DiT~\cite{dit}. By aligning the training objective with the diffusion head's capacity for \textbf{joint distribution modeling}, BitDance achieves efficient and accurate parallel prediction.

For class-conditional generation, BitDance is optimized solely using Eq.~(\ref{eq:diff_loss_patch}). For text-to-image synthesis, where the Transformer is initialized from an LLM, we additionally incorporate a standard cross-entropy loss on text tokens to preserve the model's text-understanding capabilities.

\section{Experiments}

\subsection{Scaling up Token Entropy}

We first investigate the impact of scaling token entropy on the reconstruction performance of our visual tokenizer. To this end, we evaluate three distinct configurations: (i) a 16$\times$ downsampling ratio with a codebook size of $\mathbf{2^{32}}$, (ii) a 32$\times$ downsampling ratio with a codebook size of $\mathbf{2^{128}}$, and (iii) a 32$\times$ downsampling ratio with a codebook size of $\mathbf{2^{256}}$. We utilize DataComp-1B~\cite{gadre2023datacomp} as our main training set, supplemented with high-quality face and text datasets to improve domain-specific reconstruction. Training is conducted at a 256$\times$256 resolution for 400K steps using a batch size of 1024. Following previous works~\cite{luo2024open,zhuang2025wetok}, the tokenizer adopts a pure CNN architecture, enabling generalization on various resolutions.

\begin{table}[t!]
    \centering
    \setlength{\tabcolsep}{5pt}
    \renewcommand\arraystretch{1.0}
    \caption{\textbf{Reconstruction on ImageNet 256${\times}$256 validation set.} All models are trained on general domain datasets. By scaling token entropy, our discrete tokenizer outperforms continuous VAEs, even at higher compression ratios.}
    \resizebox{0.85\linewidth}{!}{
        \begin{tabular}{@{}lc|ccc|cc@{}}
        \toprule
\multirow{2}{*}{\textbf{Method}} & 
\multirow{2}{*}{\makecell{\textbf{Tokenizer} \\ \textbf{Type}}} & 
\multirow{2}{*}{\makecell{\textbf{Downsample} \\ \textbf{Ratio}}} & 
\multirow{2}{*}{\makecell{\textbf{Codebook} \\ \textbf{Size}}} & 
\multirow{2}{*}{\makecell{\textbf{Compression} \\ \textbf{Ratio}}} & 
\multirow{2}{*}{{\textbf{PSNR}$\uparrow$}} & 
\multirow{2}{*}{{\textbf{SSIM}$\uparrow$}} \\
& & & & & & \\
        \midrule
        SD-VAE~\cite{rombach2022high} & Continuous & 8 & - & 24 & 23.54 & 0.68 \\ 
        Cosmos~\cite{agarwal2025cosmos} & Discrete& 16 & 65536 & 384 &  19.93 & 0.49 \\ 
        Show-o~\cite{xie2024show} & Discrete& 16  & 8192&473 & 21.34 & 0.59 \\ 
        LlamaGen~\cite{sun2024autoregressive} & Discrete& 16  & 16384& 439 &  20.65 & 0.54 \\
        Open-MAGVIT2~\cite{luo2024open} & Discrete& 16  & $2^{18}$& 341 &  22.70 & 0.64 \\ 
        Infinity~\cite{han2025infinity} & Discrete & 16 & $2^{32}$ & 192 & 22.70 & - \\ 
        \rowcolor{myblue!60}
        \textbf{BitDance-Tok (Ours)} & Discrete & 16 & $\mathbf{2^{32}}$ & 192 & \textbf{24.90} & \textbf{0.72} \\ 
        \midrule
        WeTok~\cite{zhuang2025wetok} & Discrete & 32 & $2^{32}$ & 768 & 20.77 & 0.55 \\
        DC-AE~\cite{chen2024deep} & Continuous & 32 & - & 48 & 24.81 & 0.69 \\ 
        DC-AE-SANA~\cite{sana} & Continuous & 32 & - & 48 & 24.72 & 0.69 \\ 
        \rowcolor{myblue!60}
        \textbf{BitDance-Tok (Ours)} & Discrete & 32 & $\mathbf{2^{128}}$ & 192 & 23.26 & 0.67 \\ 
        \rowcolor{myblue!60}
        \textbf{BitDance-Tok (Ours)} & Discrete & 32 & $\mathbf{2^{256}}$ & 96 & \textbf{25.29} & \textbf{0.74} \\ 
    \bottomrule
    \end{tabular}}
    \label{tab:rec_in1k}
\end{table}

\begin{figure}[t!]
\vspace{1em}
\centering
\hspace{-1.5em}
\begin{minipage}{0.66\linewidth}{
\includegraphics[width=0.95\textwidth]{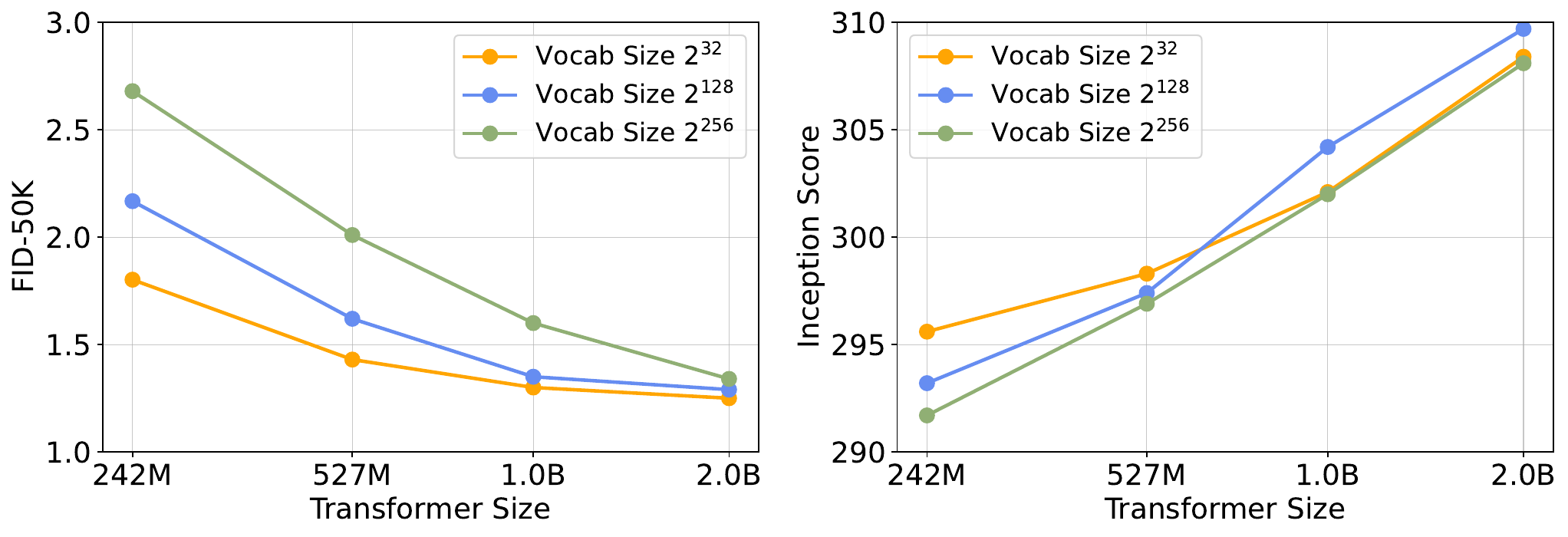}
}\end{minipage}
\hspace{-1em}
\begin{minipage}{0.34\linewidth}{
\caption{
\textbf{Generative performance across different vocabulary sizes and Transformer sizes.} For large vocabularies, small Transformers struggle to converge. Scaling the vocabulary size requires a concurrent expansion of the Transformer size.
}
\label{fig:scale_vocab}
}\end{minipage}
\end{figure}

\textbf{Scaling Token Entropy for Reconstruction.} Tab.~\ref{tab:rec_in1k} compares the performance of tokenizers under diverse downsampling and compression settings. For continuous tokenizers, we assume that latent features are stored in the commonly used bfloat16 format to calculate the compression ratio. Token entropy scaling markedly improves discrete reconstruction accuracy, bridging the gap with continuous models. The results show that with a $2^{32}$ codebook, our 16$\times$ tokenizer outperforms the continuous SD-VAE~\cite{rombach2022high}. Similarly, for 32$\times$ downsampling, scaling the codebook from $2^{128}$ to $2^{256}$ allows our discrete approach to exceed the reconstruction quality of the continuous DC-AE~\cite{chen2024deep}.

\textbf{Scaling Token Entropy and Autoregressive Model Size.} We investigate how vocabulary size influences downstream generative performance on ImageNet. To maintain uniform computational overhead across different downsampling rates, we employ class token replication to align FLOPs during training. Fig.~\ref{fig:scale_vocab} reveals that while small-scale Transformers struggle with the convergence of large vocabularies, larger models effectively leverage them to achieve superior generative quality. These findings suggest that scaling up the vocabulary is most effective when accompanied by a corresponding increase in Transformer scale.

\begin{table}[t]
\caption{\textbf{Evaluation of class-conditional image generation on ImageNet 256$\times$256.} We mark models using additional self-supervised models (e.g., DINO) in \textcolor{gray}{gray}. BitDance achieves superior performance while employing standard AR modeling with raster-order.}
\label{tab:imagenet_comparison}
\centering
\small
\setlength{\tabcolsep}{5.0pt}
\renewcommand\arraystretch{1.0}
\resizebox{0.95\linewidth}{!}{
\begin{tabular}{llr|rr|cc|cc}
\toprule
\textbf{Model} & \textbf{Type} & \textbf{Order}& \textbf{Tokenizer} & \textbf{\#Params} & \textbf{FID$\downarrow$} & \textbf{IS$\uparrow$} & \textbf{Pre.$\uparrow$} & \textbf{Rec.$\uparrow$} \\
\midrule
\multicolumn{9}{l}{\textit{Continuous Tokens}} \\
\addlinespace[2.5pt]
~ DiT-XL/2~\cite{dit}        & Diff. & - & VAE & 675M  & 2.27 & 278.2 & 0.83 & 0.57 \\
~ SiT-XL/2~\cite{ma2024sit}         & Diff. & - & VAE & 675M  & 2.06 & 277.5 & 0.83 & 0.59 \\
~ DiCo-XL~\cite{ai2025dico}         & Diff. & - & VAE & 701M  & 2.05 & 282.2 & 0.83 & 0.59 \\
~ MDTv2~\cite{gao2023mdtv2}        & Mask.+Diff. & - & VAE & 675M  & 1.58 & 314.7 & 0.79 & 0.65 \\
~ \textcolor{gray}{REPA}~\cite{repa} & \textcolor{gray}{Diff.} & \textcolor{gray}{-} & \textcolor{gray}{VAE} & \textcolor{gray}{675M}  & \textcolor{gray}{1.42} & \textcolor{gray}{305.7} & \textcolor{gray}{0.80} & \textcolor{gray}{0.65} \\
~ \textcolor{gray}{RAE}~\cite{zheng2025diffusion} & \textcolor{gray}{Diff.} & \textcolor{gray}{-} & \textcolor{gray}{RAE} & \textcolor{gray}{675M}  & \textcolor{gray}{1.13} & \textcolor{gray}{262.6} & \textcolor{gray}{0.78} & \textcolor{gray}{0.67} \\
~ MAR-B~\cite{li2024autoregressive}          & Mask.  & random & VAE & 208M  & 2.31 & 281.7 & 0.82 & 0.57 \\
~ MAR-L~\cite{li2024autoregressive}            & Mask.  & random & VAE & 479M  & 1.78 & 296.0 & 0.81 & 0.60 \\
~ MAR-H~\cite{li2024autoregressive}            & Mask.  & random& VAE  & {943M} & {1.55} & {303.7} & {0.81} & {0.62} \\
~ SphereAR-B~\cite{ke2025hyperspherical} & AR & raster& VAE & 208M  & 1.92 & 277.8 & 0.81 & 0.61 \\ 
~ SphereAR-L~\cite{ke2025hyperspherical} & AR & raster& VAE & 479M  & 1.54 & 295.9 & 0.80 & 0.63 \\ 
~ SphereAR-H~\cite{ke2025hyperspherical} & AR & raster& VAE & 943M  & 1.34 & 300.0 & 0.80 & 0.64 \\ 
~ xAR-B~\cite{xar} & AR+Diff. & raster & VAE & 172M  & 1.72 & 280.4 & 0.82 & 0.59 \\ 
~ xAR-L~\cite{xar} & AR+Diff. & raster & VAE & 608M   & 1.28 & 292.5 & 0.82 & 0.62 \\ 
~ xAR-H~\cite{xar} & AR+Diff. & raster & VAE & 1.1B  & 1.24 & 301.6 & 0.83 & 0.64 \\ 
\midrule
\multicolumn{9}{l}{\textit{Discrete Tokens}} \\
\addlinespace[2pt]
~ LlamaGen-L~\cite{sun2024autoregressive}      & AR   & raster& VQ  & 343M   & 3.07 & 256.1 & 0.83 & 0.52 \\
~ LlamaGen-XL~\cite{sun2024autoregressive}      & AR   & raster& VQ  & 775M   & 2.62 & 244.1 & 0.80 & 0.57 \\
~ LlamaGen-XXL~\cite{sun2024autoregressive}    & AR   & raster& VQ  & 1.4B   & 2.34 & 253.9 & 0.80 & 0.59 \\
~ RandAR-L~\cite{pang2025randar}      & AR   & random& VQ  & 343M   & 2.55 & 288.8 & 0.81 & 0.58 \\
~ RandAR-XL~\cite{pang2025randar}       & AR   & random& VQ  & 775M   & 2.22 & 314.2 & 0.80 & 0.60 \\
~ RandAR-XXL~\cite{pang2025randar}       & AR   & random& VQ  & 1.4B   & 2.15 & 322.0 & 0.79 & 0.62 \\
~ RAR-L~\cite{yu2025randomized}     & AR   & hybrid& VQ  & 461M   & 1.70 & 299.5 & 0.81 & 0.60 \\
~ RAR-XL~\cite{yu2025randomized}     & AR   & hybrid& VQ  & 955M   & 1.50 & 306.9 & 0.80 & 0.62 \\
~ RAR-XXL~\cite{yu2025randomized}     & AR   & hybrid& VQ  & 1.5B   & 1.48 & \textbf{326.0} & 0.80 & 0.63 \\
~ OpenMAGVIT2-XL~\cite{luo2024open}       & AR & raster & LFQ & 804M   & 2.51 & 271.7 & \textbf{0.84} & 0.54 \\
~ MAGVIT-v2~\cite{yulanguage}       & Mask. & random & LFQ & 307M & 1.78 & 319.4 & - & - \\
~ VAR-d20~\cite{tian2024visual}       & VAR & - & VQ & 600M  & 2.57 & 302.6 & 0.83 & 0.56 \\
~ VAR-d30~\cite{tian2024visual}       & VAR & - & VQ & 2B   & 1.92 & 323.1 & 0.82 & 0.59 \\
\rowcolor{myblue!60}
~ \textbf{BitDance-B-1x}       & AR & raster & LFQ & 242M   & 1.68 & 297.1 & 0.80 & 0.62 \\
\rowcolor{myblue!60}
~ \textbf{BitDance-L-1x}       & AR & raster & LFQ & 527M   & 1.31 & 300.2 & 0.80 & 0.64 \\
\rowcolor{myblue!60}
~ \textbf{BitDance-H-1x}      & AR & raster & LFQ & 1.0B   & \textbf{1.24} & 304.4 & 0.81 & \textbf{0.64} \\
\bottomrule
\end{tabular}}
\end{table}

\begin{table}[t!]
\caption{\textbf{Overall comparison with parallel image generation methods on ImageNet 256$\times$256.} Throughput is measured on one A100 with a batch size of 64 at bfloat16 precision. The suffixes "-4x" and "-16x" indicate that 4 and 16 tokens are generated in parallel per step. With only 260M parameters, our model outperforms the 1.4B SOTA model in parallel prediction, while achieving significantly faster generation speeds.}
\label{tab:imagenet_speed}
\centering
\small
\setlength{\tabcolsep}{5pt}
\renewcommand\arraystretch{1.0}
\resizebox{1.0\linewidth}{!}{
\begin{tabular}{llr|rrr|cc|cc}
\toprule
\textbf{Model} & \textbf{Type} & \textbf{Order} & \textbf{\#Params}& \textbf{Steps} & \textbf{Throughput$\uparrow$} & \textbf{FID$\downarrow$} & \textbf{IS$\uparrow$} & \textbf{Pre.$\uparrow$} & \textbf{Rec.$\uparrow$} \\
\midrule
DiT-XL/2~\cite{dit}        & Diff. &- & 675M & 250 & 1.06 img/s  & 2.27 & 278.2 & 0.83 & 0.57 \\ 
DiCo-XL~\cite{ai2025dico}        & Diff. &- & 701M & 250 & 2.62 img/s  & 2.05 & 282.2 & 0.83 & 0.59 \\ 
\midrule
MaskGIT~\cite{chang2022maskgit}  & Mask. & random  & 227M & 8 & 50.73 img/s  & 6.18 & 182.1 & 0.80  & 0.51 \\
MAR-B~\cite{li2024autoregressive}         & Mask.  & random & 208M & 256 & 1.83 img/s  & 2.31 & 281.7 & 0.82 & 0.57 \\
MAR-L~\cite{li2024autoregressive}    & Mask.& random  & 479M & 256 & 1.39 img/s  & 1.78 & 296.0 & 0.81 & 0.60 \\
\midrule
VAR-d20~\cite{tian2024visual}          & VAR  & -  & 600M & 10 & 71.31 img/s  & 2.57& 302.6& \textbf{0.83}& 0.56 \\
VAR-d24~\cite{tian2024visual}          & VAR & -   & 1.0B & 10 & 47.22 img/s  & 2.09 &312.9 & 0.82 & 0.59 \\
\midrule
PAR-L~\cite{wang2025parallelized}  & AR & hybrid & 343M &  147 & 15.01 img/s  & 3.76 &218.9& 0.81& 0.60 \\
PAR-XL~\cite{wang2025parallelized}  & AR & hybrid & 775M &  147 & 8.09 img/s  & 2.61 &259.2& 0.80& 0.62 \\
PAR-XXL~\cite{wang2025parallelized}  & AR & hybrid & 1.4B &  147 & 5.17 img/s  & 2.35 &263.2& 0.80& 0.62 \\
NAR-L~\cite{he2025neighboring}  & AR & hybrid & 372M &  31 & 40.03 img/s  & 3.06 &263.9& 0.81& 0.53 \\
NAR-XL~\cite{he2025neighboring}  & AR & hybrid & 816M &  31 & 23.12 img/s  & 2.70 &277.5& 0.81& 0.58 \\
NAR-XXL~\cite{he2025neighboring}  & AR & hybrid & 1.5B &  31 & 15.37 img/s  & 2.58 &293.5& 0.82& 0.57\\
RandAR-L~\cite{pang2025randar}  & AR & random & 343M &  88 & 25.12 img/s  & 2.55 &288.8& 0.81 &0.58\\
RandAR-XL~\cite{pang2025randar}  & AR & random & 775M &  88 & 16.01 img/s  & 2.25 &317.8& 0.80& 0.60\\
RandAR-XXL~\cite{pang2025randar}  & AR & random & 1.4B &  88 & 10.39 img/s  & 2.15& \textbf{322.0}& 0.79& 0.62\\
\rowcolor{myblue!60}
\textbf{BitDance-B-4x}       & AR & raster & 260M & 64 &  24.18 img/s   & \textbf{1.69} & 291.2 & 0.79 & \textbf{0.63} \\
\rowcolor{myblue!60}
\textbf{BitDance-B-16x}       & AR & raster & 260M & 16 & \textbf{90.26} img/s   & 1.91 & 283.8 & 0.78 & {0.62} \\
\bottomrule
\end{tabular}}
\end{table}

\begin{table}[t]
\small
\centering
\caption{\textbf{Training recipe of BitDance for text-to-image generation.} In the distillation phase, which requires only a few training steps, we transition a model with 16-token parallel prediction ($p=4$) into a model with 64-token parallel prediction ($p=8$) to achieve faster inference speeds.}
\label{tab:training_recipe}
\begin{tabular}{l|cccc}
\toprule
\textbf{Stage} & \textbf{PT} & \textbf{CT} & \textbf{SFT} & \textbf{Distill (optional)} \\
\midrule
Learning rate  & $1.0\times10^{-4}$ & $1.0\times10^{-4}$ & $2.0\times10^{-5}$ & $2.0\times10^{-5}$ \\
LR scheduler   & Constant & Constant & Constant & Constant\\
Weight decay   & 0.0 & 0.0 & 0.0 &0.0 \\
Gradient norm clip  & 1.0 & 1.0 & 1.0&1.0 \\
Optimizer       & \multicolumn{4}{c}{AdamW ($\beta_1=0.9$, $\beta_2=0.95$, $\epsilon=1.0 \times 10^{-15}$)} \\
Loss weight (text : vision)  & 0.01 : 1 & 0.01 : 1 & 0.01 : 1 & 0.01 : 1 \\
Warm-up steps   & 2000 & 1000 & 500 & 500 \\
Training steps  & 100K & 40K & 40K & 20K \\
Global batch size & 2560 & 2480 & 2320 & 1536 \\
\# Training samples & 256M & 99.2M & 92.8M & 30.7M \\
Image resolution (256px : 512px : 1024px) & 2 : 7 : 1 & 0 : 1 : 1 & 0 : 1 : 1 & 0 : 0 : 1 \\
Text token drop prob  & 0.1 & 0.1 & 0.1 & 0.1\\
\bottomrule
\end{tabular}
\end{table}

\subsection{Class-conditional Image Generation}
\subsubsection{Experimental Setup}

\textbf{Datasets and Metrics.} We follow prior studies~\cite{li2024autoregressive,tian2024visual,sun2024autoregressive} and evaluate our models on the class-conditional ImageNet-1K~\cite{deng2009imagenet} generation benchmark at a resolution of 256$\times$256. The main evaluation metric is the Fréchet Inception Distance (FID)~\cite{heusel2017gans}. Additionally, we report Inception Score (IS)~\cite{salimans2016improved} as well as Precision and Recall~\cite{kynkaanniemi2019improved} as complementary measures of generative quality. All evaluation metrics are computed using OpenAI’s TensorFlow-based evaluation toolkit~\cite{dhariwal2021diffusion}.

\textbf{Implementation Details.} Following MAR~\cite{li2024autoregressive}, we explore the scaling capabilities of three BitDance models of varying sizes. BitDance-B, -L, and -H feature 24, 32, and 40 Transformer blocks with widths of 768, 1024, and 1280, respectively. The binary diffusion head maintains a consistent width, while the number of blocks is set at 6, 8, and 12. Furthermore, our model nomenclature incorporates the patch size $p$ used in next-patch diffusion modeling. For instance, BitDance-B-4x denotes a patch size of $p=2$, where each AR step generates $p^2=4$ tokens. Following previous works~\cite{li2024autoregressive,ke2025hyperspherical,tian2024visual}, we conduct our experiments using a 16$\times$ downsampling binary tokenizer trained on ImageNet. The autoregressive model is trained for 800 epochs using the AdamW~\cite{loshchilov2017decoupled} optimizer with $\beta=(0.9,0.95)$ and a weight decay of 0.05. We employ a batch size of 1024 and a cosine learning rate schedule with 20K warmup steps. Additionally, an Exponential Moving Average (EMA) decay rate of 0.9999 is applied.

\subsubsection{Main Results}

We begin by exploring the scalability of BitDance when $p=1$. Tab.~\ref{tab:imagenet_comparison} indicates a consistent performance gain as the model size grows. Our largest variant, BitDance-H-1x (1B parameters), reaches an FID of 1.24, outperforming prior AR baselines even under a standard raster-scan causal framework. Such results underscore the superior modeling proficiency inherent in our approach.

We also explore parallel generation using BitDance-B as the base model for $p=2$ and $4$. Tab.~\ref{tab:imagenet_speed} reveals that our BitDance-B-4x outperforms the \textbf{1.4B}-parameter state-of-the-art parallel AR baseline RandAR-XXL~\cite{pang2025randar} by roughly \textbf{0.5} FID, despite having only \textbf{260M} parameters. These results provide strong evidence that the binary diffusion head effectively captures the joint distribution of parallel tokens, enabling both parameter efficiency and high generative quality.

\begin{table}[t!]
\caption{\textbf{Evaluation of text-to-image generation on DPG-Bench~\cite{hu2024ella}.} We mark the best and second-best performance among autoregressive models in \textbf{bold} and \underline{underline}, respectively.}
\label{tab:dpg_only}
\centering
\small
\setlength{\tabcolsep}{5pt}
\renewcommand\arraystretch{1.0}
\resizebox{0.8\linewidth}{!}{
\begin{tabular}{lcccccc}
\toprule
\textbf{Model} & \textbf{Global} & \textbf{Entity} & \textbf{Attribute} & \textbf{Relation} & \textbf{Other} & \textbf{Overall$\uparrow$} \\
\midrule
\multicolumn{7}{l}{\textit{Proprietary Models}} \\
~ GPT Image 1~\cite{openai2025gptimage} & 88.89 & 88.94 & 89.84 & 92.63 & 90.96 & 85.15 \\
~ Seedream 3.0~\cite{gao2025seedream} & 94.31 & 92.65 & 91.36 & 92.78 & 88.24 & 88.27 \\
\midrule
\multicolumn{7}{l}{\textit{Diffusion Models}} \\
~ PixArt-$\alpha$~\cite{chen2023pixart} & 86.89 & 82.89 & 88.94 & 86.59 & 87.68 & 80.54 \\
~ FLUX.1-Dev~\cite{flux} & 74.35 & 90.00 & 88.96 & 90.87 & 88.33 & 83.84 \\
~ SD3 Medium~\cite{esser2024sd3} & 87.90 & 91.01 & 88.83 & 80.70 & 88.68 & 84.08 \\
~ Z-Image-Turbo~\cite{team2025zimage} & 91.29 & 89.59 & 90.14 & 92.16 & 88.68 & 84.86 \\
~ BAGEL~\cite{deng2025bagel} & - & - & - & - & - & 85.07 \\
~ HiDream-I1-Full~\cite{cai2025hidream} & 76.44 & 90.22 & 89.48 & 93.74 & 91.83 & 85.89 \\
~ Lumina-Image-2.0~\cite{qin2025lumina} & - & 91.97 & 90.20 & 94.85 & - & 87.20 \\
~ Z-Image~\cite{team2025zimage} & 93.39 & 91.22 & 93.16 & 92.22 & 91.52 & 88.14 \\
~ Qwen-Image~\cite{qwenimage} & 91.32 & 91.56 & 92.02 & 94.31 & 92.73 & 88.32 \\
\midrule
\multicolumn{7}{l}{\textit{Autoregressive Models}} \\
~ Emu3-Gen~\cite{wang2024emu3} & 85.21 & 86.68 & 86.84 & 90.22 & 83.15 & 80.60 \\
~ Infinity~\cite{han2025infinity} & \textbf{93.11} & - & - & 90.76 & - & 83.46 \\
~ Janus-Pro~\cite{chen2025janus} & 86.90 & 88.90 & \underline{89.40} & 89.32 & 89.48 & 84.19 \\
~ Tar~\cite{han2025tar} & 83.98 & 88.62 & 88.05 & \textbf{93.98} & 84.86 & 84.19 \\
~ NextStep-1~\cite{team2025nextstep} & - & - & - & - & - & \underline{85.28} \\
~ GLM-Image~\cite{glmimage} & 87.74 & \underline{90.25} & 89.08 & \underline{92.15} & \underline{90.17} & 84.78 \\
\rowcolor{myblue!60}
~ \textbf{BitDance} & \underline{89.53} & \textbf{93.76} & \textbf{92.47} & 91.81 & \textbf{90.26} & \textbf{88.28} \\
\bottomrule
\end{tabular}
}
\end{table}

\begin{table}[t]
\caption{\textbf{Evaluation of text-to-image generation on GenEval~\cite{ghosh2023geneval}.} We mark the best and second-best performance among autoregressive models in \textbf{bold} and \underline{underline}, respectively.}
\label{tab:geneval_only}
\centering
\small
\setlength{\tabcolsep}{5pt}
\renewcommand\arraystretch{1.0}
\resizebox{0.85\linewidth}{!}{
\begin{tabular}{lccccccc}
\toprule
\textbf{Model} &\textbf{Single Obj.} & \textbf{Two Obj.} & \textbf{Count} & \textbf{Colors} & \textbf{Pos.} & \textbf{Color Attri.} & \textbf{Overall$\uparrow$} \\
\midrule
\multicolumn{7}{l}{\textit{Proprietary Models}} \\
~ GPT Image 1~\cite{openai2025gptimage}   & 0.99  & 0.92 & 0.85 & 0.92 & 0.75 & 0.61 & 0.84 \\
~ Seedream 3.0~\cite{gao2025seedream}   & 0.99      & 0.96 & 0.91 & 0.93 & 0.47 & 0.80 & 0.84 \\
\midrule
\multicolumn{7}{l}{\textit{Diffusion Models}} \\
~ PixArt-$\alpha$~\cite{chen2023pixart}  & 0.98     & 0.50 & 0.44 & 0.80 & 0.08 & 0.07 & 0.48 \\
~ SD3 Medium~\cite{esser2024sd3}        & 0.98      & 0.74 & 0.63 & 0.67 & 0.34 & 0.36 & 0.62 \\
~ JanusFlow~\cite{ma2024janusflow}      & 0.97      & 0.59 & 0.45 & 0.83 & 0.53 & 0.42 & 0.63 \\
~ FLUX.1-Dev~\cite{flux}            & 0.98          & 0.81 & 0.74 & 0.79 & 0.22 & 0.45 & 0.66 \\
~ SD3.5-Large~\cite{esser2024sd3}        & 0.98     & 0.89 & 0.73 & 0.83 & 0.34 & 0.47 & 0.71 \\
~ Lumina-Image-2.0~\cite{qin2025lumina}   & -    & 0.87 & 0.67 & -    & -    & 0.62 & 0.73 \\
~ Show-o2~\cite{xie2025show}             & 1.00      & 0.87 & 0.58 & 0.92 & 0.52 & 0.62 & 0.76 \\
~ Z-Image-Turbo~\cite{team2025zimage} & 1.00      & 0.95 & 0.77 & 0.89 & 0.65 & 0.68 & 0.82 \\
~ HiDream-I1-Full~\cite{cai2025hidream}   & 1.00    & 0.98 & 0.79 & 0.91 & 0.60 & 0.72 & 0.83 \\
~ Z-Image~\cite{team2025zimage}           & 1.00    & 0.94 & 0.78 & 0.93 & 0.62 & 0.77 & 0.84 \\
~ Qwen-Image~\cite{qwenimage}          & 0.99      & 0.92 & 0.89 & 0.88 & 0.76 & 0.77 & 0.87 \\
~ BAGEL~\cite{deng2025bagel}             & 0.98   & 0.95 & 0.84 & 0.95 & 0.78 & 0.77 & 0.88 \\
\midrule
\multicolumn{7}{l}{\textit{Autoregressive Models}} \\
~ Emu3-Gen~\cite{wang2024emu3}            & 0.98    & 0.71 & 0.34 & 0.81 & 0.17 & 0.21 & 0.54 \\
~ Infinity~\cite{han2025infinity}       & -   & 0.85 & -    & -    & 0.49 & 0.57 & 0.73 \\
~ Janus-Pro~\cite{chen2025janus}        & \underline{0.99}      & 0.89 & 0.59 & \underline{0.90} & \underline{0.79} & \underline{0.66} & 0.80 \\
~ Tar~\cite{han2025tar}         & 0.98 & \underline{0.92} & \textbf{0.83} & 0.85 & \textbf{0.80} & 0.65 & \underline{0.84} \\
~ NextStep-1~\cite{team2025nextstep}   & -       & -    & -    & -    & -    & -    & 0.73 \\
\rowcolor{myblue!60}
~ \textbf{BitDance}                 & \textbf{1.00}   & \textbf{0.96} & \underline{0.71} & \textbf{0.95} & 0.72 & \textbf{0.83} & \textbf{0.86} \\
\bottomrule
\end{tabular}
}
\end{table}

\begin{table}[t]
\caption{\textbf{Evaluation of text-to-image generation on OneIG-EN~\cite{chang2025oneig}.} We mark the best and second-best performance among autoregressive models in \textbf{bold} and \underline{underline}, respectively.}
\label{tab:oneig_en}
\centering
\small
\setlength{\tabcolsep}{5pt}
\renewcommand\arraystretch{1.0}
\resizebox{0.82\linewidth}{!}{
\begin{tabular}{lcccccc}
\toprule
\textbf{Model} & \textbf{Alignment} & \textbf{Text} & \textbf{Reasoning} & \textbf{Style} & \textbf{Diversity} & \textbf{Overall$\uparrow$} \\
\midrule
\multicolumn{7}{l}{\textit{Proprietary Models}} \\
~ Imagen 4~\cite{google2025imagen4}         & 0.857 & 0.805 & 0.338 & 0.377 & 0.199 & 0.515 \\
~ Seedream 3.0~\cite{gao2025seedream}       & 0.818 & 0.865 & 0.275 & 0.413 & 0.277 & 0.530 \\
~ GPT Image 1~\cite{openai2025gptimage}     & 0.851 & 0.857 & 0.345 & 0.462 & 0.151 & 0.533 \\
\midrule
\multicolumn{7}{l}{\textit{Diffusion Models}} \\
~ Show-o2~\cite{xie2025show}                & 0.817 & 0.002 & 0.226 & 0.317 & 0.177 & 0.308 \\
~ SANA-1.5~\cite{xie2025sana}               & 0.765 & 0.069 & 0.217 & 0.401 & 0.216 & 0.334 \\
~ BAGEL~\cite{deng2025bagel}                & 0.769 & 0.244 & 0.173 & 0.367 & 0.251 & 0.361 \\
~ FLUX.1-Dev~\cite{flux}                    & 0.786 & 0.523 & 0.253 & 0.368 & 0.238 & 0.434 \\
~ OmniGen2~\cite{wu2025omnigen2}            & 0.804 & 0.680 & 0.271 & 0.377 & 0.242 & 0.475 \\
~ HiDream-I1-Full~\cite{cai2025hidream}     & 0.829 & 0.707 & 0.317 & 0.347 & 0.186 & 0.477 \\
~ Z-Image-Turbo~\cite{team2025zimage}       & 0.840 & 0.994 & 0.298 & 0.368 & 0.139 & 0.528 \\
~ Qwen-Image~\cite{qwenimage}               & 0.882 & 0.891 & 0.306 & 0.418 & 0.197 & 0.539 \\
~ Z-Image~\cite{team2025zimage}             & 0.881 & 0.987 & 0.280 & 0.387 & 0.194 & 0.546 \\
\midrule
\multicolumn{7}{l}{\textit{Autoregressive Models}} \\
~ Janus-Pro~\cite{chen2025janus}            & 0.553 & 0.001 & 0.139 & 0.276 & \textbf{0.365} & 0.267 \\
~ NextStep-1~\cite{team2025nextstep}        & \underline{0.826} & 0.507 & 0.224 & 0.332 & 0.199 & 0.418 \\
~ GLM-Image~\cite{glmimage}        & 0.805&	\textbf{0.969}&	\textbf{0.298}&	\underline{0.353}&	\underline{0.213}&	\underline{0.528} \\
\rowcolor{myblue!60}
~ \textbf{BitDance}                  & \textbf{0.853} & \underline{0.937} & \underline{0.297} & \textbf{0.395} & 0.177 & \textbf{0.532} \\
\bottomrule
\end{tabular}
}
\end{table}

\begin{table}[t!]
\caption{\textbf{Evaluation of text-to-image generation on OneIG-ZH~\cite{chang2025oneig}.} We mark the best and second-best performance among autoregressive models in \textbf{bold} and \underline{underline}, respectively.}
\label{tab:oneig_zh}
\centering
\small
\setlength{\tabcolsep}{5pt}
\renewcommand\arraystretch{1.0}
\resizebox{0.82\linewidth}{!}{
\begin{tabular}{lcccccc}
\toprule
\textbf{Model} & \textbf{Alignment} & \textbf{Text} & \textbf{Reasoning } & \textbf{Style} & \textbf{Diversity} & \textbf{Overall$\uparrow$} \\
\midrule
\multicolumn{7}{l}{\textit{Proprietary Models}} \\
~ Kolors 2.0~\cite{kuaishou2025kolors}      & 0.738 & 0.502 & 0.226 & 0.331 & 0.333 & 0.426 \\
~ GPT Image 1~\cite{openai2025gptimage}     & 0.812 & 0.650 & 0.300 & 0.449 & 0.159 & 0.474 \\
~ Seedream 3.0~\cite{gao2025seedream}       & 0.793 & 0.928 & 0.281 & 0.397 & 0.243 & 0.528 \\
\midrule
\multicolumn{7}{l}{\textit{Diffusion Models}} \\
~ HiDream-I1-Full~\cite{cai2025hidream}     & 0.620 & 0.205 & 0.256 & 0.304 & 0.300 & 0.337 \\
~ CogView4~\cite{zheng2024cogview3}         & 0.700 & 0.193 & 0.236 & 0.348 & 0.214 & 0.338 \\
~ BAGEL~\cite{deng2025bagel}                & 0.672 & 0.365 & 0.186 & 0.357 & 0.268 & 0.370 \\
~ Z-Image-Turbo~\cite{team2025zimage}       & 0.782 & 0.982 & 0.276 & 0.361 & 0.134 & 0.507 \\
~ Qwen-Image~\cite{qwenimage}               & 0.825 & 0.963 & 0.267 & 0.405 & 0.279 & 0.548 \\
~ Z-Image~\cite{team2025zimage}             & 0.793 & 0.988 & 0.266 & 0.386 & 0.243 & 0.535 \\
\midrule
\multicolumn{7}{l}{\textit{Autoregressive Models}} \\
~ Janus-Pro~\cite{chen2025janus}            & 0.324 & 0.148 & 0.104 & 0.264 & \textbf{0.358} & 0.240 \\
~ GLM-Image~\cite{glmimage} &	\underline{0.738}&	\textbf{0.976}&	\textbf{0.284}&	\underline{0.335}&	\underline{0.221}&	\underline{0.511} \\
\rowcolor{myblue!60}
~ \textbf{BitDance}                  & \textbf{0.786} & \underline{0.961} & \underline{0.276} & \textbf{0.376} & 0.159 & \textbf{0.512} \\
\bottomrule
\end{tabular}
}
\end{table}

\begin{table}[t!]
\centering
\caption{\textbf{Evaluation of text-to-image generation on TIIF Bench testmini~\cite{wei2025tiif}.} We mark the best and second-best performance among autoregressive models in \textbf{bold} and \underline{underline}, respectively.}
\label{tab:tiif}
\centering
\small
\setlength{\tabcolsep}{3pt}
\renewcommand\arraystretch{1.0}
\resizebox{1.0\linewidth}{!}{
\begin{tabular}{l | cc | cc cc cc cc cc cc cc cc cc cc cc}
\toprule
\multirow{3}{*}{\textbf{Model}} & \multicolumn{2}{c|}{\multirow{2}{*}{\textbf{Overall}$\uparrow$}} & \multicolumn{8}{c}{\textbf{Basic Following}} & \multicolumn{12}{c}{\textbf{Advanced Following}} & \multicolumn{2}{c}{\textbf{Designer}} \\
\cmidrule(lr){4-11} \cmidrule(lr){12-23} \cmidrule(lr){24-25}
& & & \multicolumn{2}{c}{Avg} & \multicolumn{2}{c}{Attribute} & \multicolumn{2}{c}{Relation} & \multicolumn{2}{c}{Reasoning} & \multicolumn{2}{c}{Avg} & \multicolumn{2}{c}{Attr.+Rela.} & \multicolumn{2}{c}{Attr.+Reas.} & \multicolumn{2}{c}{Rela.+Reas.} & \multicolumn{2}{c}{Style} & \multicolumn{2}{c}{Text} & \multicolumn{2}{c}{Real World} \\
& short & long & short & long & short & long & short & long & short & long & short & long & short & long & short & long & short & long & short & long & short & long & short & long \\
\midrule
\multicolumn{25}{l}{\textit{Proprietary Models}} \\
Midjourney V7~\cite{midjourneyV7} & 68.74 & 65.69 & 77.41 & 76.00 & 77.58 & 81.83 & 82.07 & 76.82 & 72.57 & 69.32 & 64.66 & 60.53 & 67.20 & 62.70 & 81.22 & 71.59 & 60.72 & 64.59 & 83.33 & 80.00 & 24.83 & 20.83 & 68.83 & 63.61 \\
DALL-E 3~\cite{betker2023dalle3} & 74.96 & 70.81 & 78.72 & 78.50 & 79.50 & 79.83 & 80.82 & 78.82 & 75.82 & 76.82 & 73.39 & 67.27 & 73.45 & 67.20 & 72.01 & 71.34 & 63.59 & 60.72 & 89.66 & 86.67 & 66.83 & 54.83 & 72.93 & 60.99 \\
Seedream 3.0~\cite{gao2025seedream} & 86.02 & 84.31 & 87.07 & 84.93 & 90.50 & 90.00 & {89.85} & 85.94 & 80.86 & 78.86 & 79.16 & 80.60 & 79.76 & 81.82 & 77.23 & 78.85 & 75.64 & 78.64 & {100.0} & 93.33 & {97.17} & 87.78 & 83.21 & 83.58 \\
GPT Image 1~\cite{openai2025gptimage} & {89.15} & {88.29} & {90.75} & {89.66} & {91.33} & 87.08 & 84.57 & 84.57 & {96.32} & {97.32} & {88.55} & {88.35} & {87.07} & {89.44} & {87.22} & {83.96} & {85.59} & {83.21} & 90.00 & 93.33 & 89.83 & 86.83 & 89.73 & {93.46} \\
\midrule
\multicolumn{25}{l}{\textit{Diffusion Models}} \\
Lumina-Next~\cite{zhuo2024lumina} & 50.93 & 52.46 & 64.58 & 66.08 & 56.83 & 59.33 & 67.57 & 71.82 & 69.32 & 67.07 & 44.75 & 45.63 & 51.44 & 43.20 & 51.09 & 59.72 & 44.72 & 54.46 & 70.00 & 66.67 & 0.00 & 0.83 & 47.56 & 49.05 \\
Hunyuan-DiT~\cite{li2024hunyuandit} & 51.38 & 53.28 & 69.33 & 69.00 & 65.83 & 69.83 & 78.07 & 73.82 & 64.07 & 63.32 & 42.62 & 45.45 & 50.20 & 41.57 & 59.22 & 61.84 & 47.84 & 51.09 & 56.67 & 73.33 & 0.00 & 0.83 & 40.10 & 44.20 \\
PixArt-$\Sigma$~\cite{chen2024pixart} & 62.00 & 58.12 & 70.66 & 75.25 & 69.33 & 78.83 & 75.07 & 77.32 & 67.57 & 69.57 & 57.65 & 49.50 & 65.20 & 56.57 & 66.96 & 61.72 & 66.59 & 54.59 & 83.33 & 70.00 & 1.83 & 1.83 & 62.11 & 52.41 \\
SANA 1.5~\cite{xie2025sana} & 67.15 & 65.73 & 79.66 & 77.08 & 79.83 & 77.83 & 85.57 & 83.57 & 73.57 & 69.82 & 61.50 & 60.67 & 65.32 & 56.57 & 69.96 & 73.09 & 62.96 & 65.84 & 80.00 & 80.00 & 17.83 & 15.83 & 71.07 & 68.83 \\
SD 3~\cite{esser2024sd3} & 67.46 & 66.09 & 78.32 & 77.75 & 83.33 & 79.83 & 82.07 & 78.82 & 71.07 & 74.07 & 61.46 & 59.56 & 61.07 & 64.07 & 68.84 & 70.34 & 50.96 & 57.84 & 66.67 & 76.67 & 59.83 & 20.83 & 63.23 & 67.34 \\
FLUX.1-dev~\cite{flux} & 71.09 & 71.78 & 83.12 & 78.65 & 87.05 & 83.17 & 87.25 & 80.39 & 75.01 & 72.39 & 65.79 & 68.54 & 67.07 & 73.69 & 73.84 & 73.34 & 69.09 & 71.59 & 66.67 & 66.67 & 43.83 & 52.83 & 70.72 & 71.47 \\
{Z-Image-Turbo}~\cite{team2025zimage} & 77.73 & 80.05 & 81.85 & 81.59 & 86.50 & 87.00 & 82.88 & 79.99 & 76.17 & 77.77 & 68.32 & 74.69 & 72.04 & 75.24 & 60.22 & 73.33 & 68.90 & 71.92 & 83.33 & 93.33 & 83.71 & 84.62 & 85.82 & 77.24 \\
{Z-Image}~\cite{team2025zimage} & 80.20 & 83.04 & 78.36 & 82.79 & 79.50 & 86.50 & 80.45 & 79.94 & 75.13 & 81.94 & 72.89 & 77.02 & 72.91 & 77.56 & 66.99 & 73.82 & 73.89 & 75.62 & 90.00 & 93.33 & 94.84 & {93.21} & 88.06 & 85.45 \\
Qwen-Image~\cite{qwenimage} & 86.14 & 86.83 & 90.18 & 87.22 & 90.50 & {91.50} & 88.22 & {90.78} & 79.81 & 79.38 & 79.30 & 80.88 & 79.21 & 78.94 & 78.85 & 81.69 & 75.57 & 78.59 & {100.0} & {100.0} & 92.76 & 89.14 & {90.30} & 91.42 \\
\midrule
\multicolumn{25}{l}{\textit{Autoregressive Models}} \\
LightGen~\cite{wu2025lightgen} 
& 53.22 & 43.41 & 66.58 & 47.91 & 55.83 & 47.33 & 74.82 & 45.82 & 69.07 & 50.57 
& 46.74 & 41.53 & 62.44 & 40.82 & 61.71 & 50.47 & 50.34 & 45.34 & 53.33 & 53.33 
& 0.00 & 6.83 & 50.92 & 50.55 \\
Infinity~\cite{han2025infinity} 
& 62.07 & 62.32 & 73.08 & 75.41 & 74.33 & 76.83 & 72.82 & \underline{77.57} & 72.07 & 71.82 
& 56.64 & 54.98 & 60.44 & 55.57 & \textbf{74.22} & 64.71 & 60.22 & 59.71 & \underline{80.00} & \underline{73.33} 
& 10.83 & 23.83 & 54.28 & 56.89 \\
Janus-Pro~\cite{chen2025januspro} 
& \underline{66.50} & \underline{65.02} & \textbf{79.33} & \underline{78.25} & \underline{79.33} & \underline{82.33} & \textbf{78.32} & 73.32 & \textbf{80.32} & \textbf{79.07} 
& \underline{59.71} & \underline{58.82} & \underline{66.07} & \underline{56.20} & \underline{70.46} & \textbf{70.84} & \underline{67.22} & \underline{59.97} & 60.00 & 70.00 
& \underline{28.83} & \underline{33.83} & \underline{65.84} & \underline{60.25} \\
GLM-Image~\cite{glmimage} 
& \textbf{81.01} & \textbf{81.02} & - & - & - & - & - & - & - & - 
& - & - & - & - & - & - & - & - & - & - 
& - & - & - & - \\
\rowcolor{myblue!60}
\textbf{BitDance} 
& \underline{79.64} & \underline{78.12} & \underline{78.79} & \textbf{80.44} & \textbf{83.50} & \textbf{87.50} & \underline{77.22} & \textbf{77.62} & \underline{75.64} & \underline{76.21} 
& \textbf{72.19} & \textbf{71.66} & \textbf{72.89} & \textbf{78.88} & 68.06 & \underline{66.10} & \textbf{70.63} & \textbf{67.21} & \textbf{96.67} & \textbf{90.00} 
& \textbf{87.78} & \textbf{75.57} & \textbf{84.33} & \textbf{83.96} \\

\bottomrule
\end{tabular}
}
\end{table}

\begin{figure}[t]
\vspace{1mm}
    \centering
    \hspace{-1.5em}
    \begin{minipage}{0.56\linewidth}
        \centering
        \includegraphics[width=0.9\linewidth]{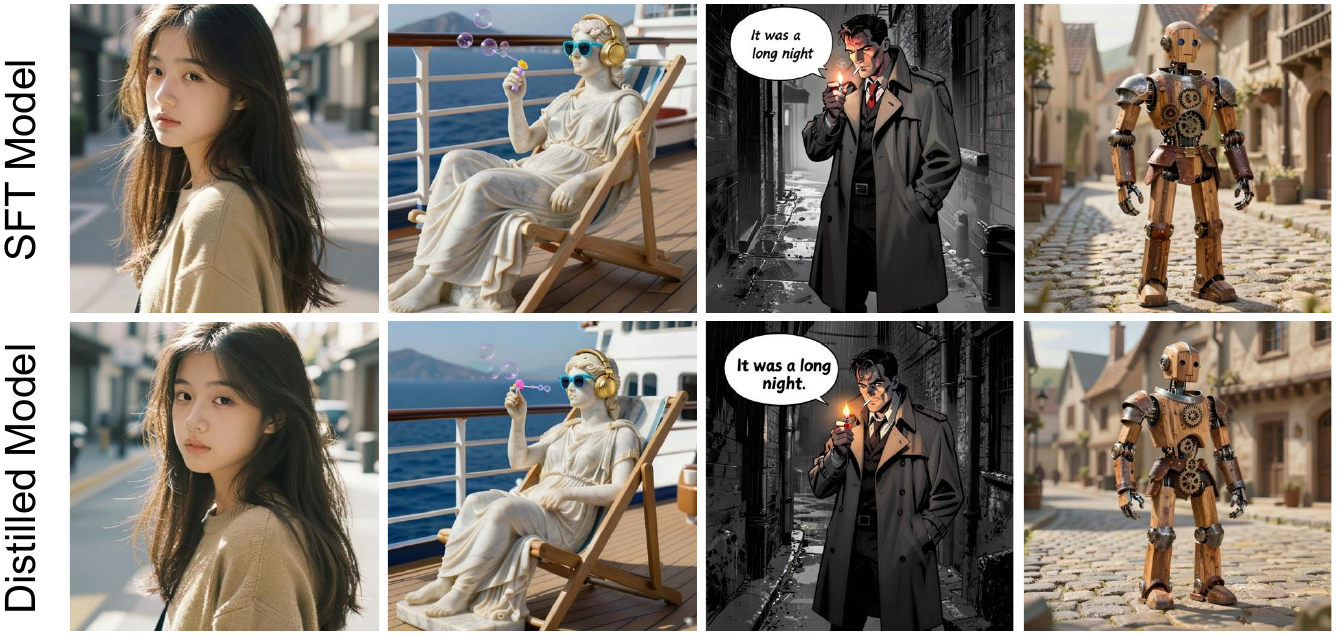}
        \caption{\textbf{Images generated by the SFT and distilled models.}}
        \label{fig:sft_vs_distill}
    \end{minipage}
    \begin{minipage}{0.44\linewidth}
        \centering
        \small  

        \captionof{table}{\textbf{Quantitative results of the SFT and distilled models.} \#Tokens denotes the number of predicted tokens in one decoding step.}
        \label{tab:sft_vs_distill}

        \resizebox{\linewidth}{!}{%
            \begin{tabular}{lccc}
                \toprule
               \textbf{Model} &\textbf{\#Tokens} & \textbf{DPG-Bench}  & \textbf{GenEval} \\
                \midrule
                SFT     & 16  & 88.28  & \textbf{0.86} \\
                Distilled & \textbf{64} &\textbf{88.30}  & 0.85 \\
                \bottomrule
            \end{tabular}
        }
    \end{minipage}

\end{figure}

\begin{table}[t!]
    \centering
    \caption{\textbf{Efficiency comparison of text-to-image generation at 1024×1024 resolution.} Latency is measured using a single H100 GPU with bfloat16 precision.}
    \label{tab:t2i_speed}
    \resizebox{0.65\linewidth}{!}{
    \begin{tabular}{lccc}
        \toprule
        \textbf{Model} & \textbf{Type} & \textbf{\#Params (B)}  & \textbf{Latency-1024 (s)} \\
        \midrule
        BAGEL~\cite{deng2025bagel}       & Diffusion  & 7 & 23.1 \\
        Qwen-Image~\cite{qwenimage}       & Diffusion  & 20 & 20.3 \\
        Z-Image~\cite{team2025zimage}       & Diffusion  & 6 & 21.1 \\
        NextStep-1~\cite{team2025nextstep}       & Autoregressive  & 14 & 402 \\
        GLM-Image~\cite{glmimage}       & Autoregressive  & 16 & 53.2 \\
        \rowcolor{myblue!60}
        \textbf{BitDance} & Autoregressive & 14 & 12.4 \\
        \bottomrule
    \end{tabular}}
\end{table}

\subsection{Text-to-image Generation}
\subsubsection{Training Details}
\textbf{Model Design.} To scale up BitDance on the text-to-image generation task, we choose the pretrained Qwen3-14B~\cite{yang2025qwen3} as the base AR model. Following previous works~\cite{team2025nextstep,wang2024emu3}, the AR model serves as both the text encoder and image generator. Our primary goal is to enhance it with image generation capability using high-quality image-text pairs. To balance reconstruction quality and generative convergence speed, we choose the tokenizer with 16x downsampling ratio for text-to-image generation. To achieve more efficient high-resolution generation, we set the patch size $p$ to 4 and enable parallel prediction of $p^2=16$ tokens per step. For positional encoding, in addition to the 1D RoPE~\cite{su2024roformer} used in LLM, we incorporate 2D sinusoidal positional embeddings to enhance the model's spatial awareness of images.

Notably, to support \textbf{varying image resolution and aspect ratios}, we add two resolution tokens \texttt{[res\_i]} and \texttt{[res\_j]} at the beginning of visual tokens:
\begin{center}
\texttt{[bos], \{text tokens\}, [boi], [res\_i], [res\_j], \{visual tokens\}, [eoi], [eos]},
\end{center}
where $i$ and $j$ indicate the number of visual tokens on the height and width dimensions. In this way, we can decide how many tokens (i.e., $i \times j$) to be decoded during inference.

\textbf{Training Recipe.} We adopt a three-stage training pipeline, including \textbf{pre-training (PT)}, \textbf{continued training (CT)} and \textbf{supervised fine-tuning (SFT)}. Our PT and CT data is collected from open-sourced datasets such as LAION~\cite{schuhmann2022laion}. The SFT data is collected from both open-source datasets and a small number of images generated by other text-to-image models such as Seedream~\cite{gao2025seedream} and Z-Image-Turbo~\cite{team2025zimage}, which help our model to quickly adapt to the distribution of high-quality images and achieve strong visual fidelity with substantially fewer training samples. Besides, we also preserve a small amount of image-to-text data to maintain the text understanding ability of LLM.

To further accelerate model inference, we introduce an additional \textbf{distillation stage} that transitions the SFT model from 16-token parallel prediction to 64-token parallel prediction. Specifically, we inherit the weights from the SFT model and fine-tune it using a small amount of high-quality data. We find that the model can adapt to predicting more tokens in parallel with only a few training steps.

The training parameters are listed in Tab.~\ref{tab:training_recipe}. Different from prior works, we adopt a \textbf{mixed-resolution training strategy} during pre-training, which we find crucial for training stability. Specifically, we use 512px as the primary training resolution, while jointly incorporating 256px images to improve training throughput and 1024px images to enhance stability and robustness at higher resolutions. This mixed-resolution setup enables stable optimization while maintaining both efficiency and high-resolution generalization. During the continued training stage, we increase the proportion of 1024px images to enhance the model's performance in high-resolution image generation.

For all training stages, we use the AdamW~\cite{loshchilov2017decoupled} optimizer with $\beta=(0.9,0.95)$ and a constant learning rate. The visual tokenizer is frozen, while all other model parameters remain trainable. The training loss consists of two parts: a vision loss from the binary diffusion head and a text loss (cross-entropy loss) from the AR model. We set the relative weighting of vision and text loss as 1 : 0.01, which allows the AR model to maintain fundamental text encoding capabilities while keeping the primary focus on image generation. To enable classifier-free guidance~\cite{ho2022classifier} and improve model robustness, we randomly drop text tokens with a predefined token dropout probability of 0.1 during training. This encourages the model to rely on both text and image cues and prevents overfitting to the textual input.

\subsubsection{Main Results}
We conduct a comprehensive evaluation of BitDance on text-to-image generation, covering capabilities such as prompt following, text rendering, and reasoning. We compare it against state-of-the-art proprietary models (e.g., Seeddream~\cite{gao2025seedream}, GPT Image 1~\cite{openai2025gptimage}), diffusion models (e.g., Qwen-Image~\cite{qwenimage}, Z-Image~\cite{team2025zimage}), and autoregressive models (e.g., GLM-Image~\cite{glmimage}, NextStep-1~\cite{team2025nextstep}). The results demonstrate that BitDance achieves state-of-the-art performance among autoregressive models and is comparable to leading proprietary and diffusion models.

As shown in Tab.~\ref{tab:dpg_only}, Tab.~\ref{tab:geneval_only} and Tab.~\ref{tab:tiif}, BitDance demonstrates strong prompt-following capabilities.  On GenEval~\cite{ghosh2023geneval} and DPG-Bench~\cite{hu2024ella}, BitDance outperforms the majority of existing open-source models across most evaluation dimensions. BitDance achieves an overall GenEval score of 0.86 and DPG-Bench score of 88.28, ranking among the top-performing methods. On the TIIF benchmark (testmini)~\cite{wei2025tiif}, BitDance achieves the second-best overall performance among autoregressive models, demonstrating its strong capability in executing complex and diverse user instructions. Importantly, these competitive results are achieved with \textbf{significantly fewer training data.} BitDance is trained on fewer than 450M image–text pairs, which is orders of magnitude smaller than the billion- to multi-billion-scale datasets typically used by current leading commercial models. Despite this substantial data disadvantage, BitDance is able to match or closely approach the performance of such proprietary systems on multiple benchmarks.

Results on OneIG Bench~\cite{chang2025oneig} further validate the data efficiency and generalization ability of BitDance. As shown in Tab.\ref{tab:oneig_en} and Tab.~\ref{tab:oneig_zh}, on both OneIG-EN and OneIG-ZH, BitDance achieves competitive average scores, narrowing the gap with state-of-the-art commercial models, particularly in text fidelity and alignment. Taken together, these results highlight that BitDance
effectively bridges the performance gap between open-source and proprietary models while relying on dramatically less training data, underscoring the effectiveness and scalability of our method.

Furthermore, we provide qualitative and quantitative results for both the SFT and distilled models in Fig.~\ref{fig:sft_vs_distill} and Tab.~\ref{tab:sft_vs_distill}. These results demonstrate that the distilled model maintains excellent generation quality while achieving faster inference speeds. In Tab.~\ref{tab:t2i_speed}, we present an efficiency comparison of various text-to-image models generating 1024×1024 images. Compared to both diffusion and autoregressive models, our distilled model achieves faster inference speeds.

\begin{table}[t]
    \centering
    \begin{minipage}{0.30\textwidth}
        \centering
        \caption{\textbf{Different visual tokenizers for AR generation} ($p=1$). }
        \label{tab:abla_vae}
        \small
        \begin{tabular}{lcc}
            \toprule
            \textbf{Tokenizer} & \textbf{FID}$\downarrow$  & \textbf{IS}$\uparrow$ \\
            \midrule
            MAR's VAE~\cite{li2024autoregressive} & 3.16  & 289.9 \\
            VA-VAE~\cite{yao2025reconstruction} & 4.84  & 273.7 \\
            \rowcolor{myblue!60}
            \textbf{BitDance-Tok} & \textbf{1.79}  & \textbf{290.5} \\
            \bottomrule
        \end{tabular}
    \end{minipage}
    \hfill 
    \begin{minipage}{0.32\textwidth}
        \centering
        \caption{\textbf{Comparison of different sampling heads} ($p=1$).}
        \label{tab:abla_head}
        \small 
        \begin{tabular}{lcc}
            \toprule
            \textbf{Sampling Head} & \textbf{FID}$\downarrow$  & \textbf{IS}$\uparrow$ \\
            \midrule
            Token Cls Head       & OOM  & OOM \\
            Bit-wise Cls Head  & 8.37  & 174.5 \\
            \rowcolor{myblue!60}
            \textbf{Binary Diff Head} & \textbf{1.79} & \textbf{290.5} \\
            \bottomrule
        \end{tabular}
    \end{minipage}
    \hfill
    \begin{minipage}{0.34\textwidth}
        \centering
        \caption{\textbf{Ablation study on design of next-patch diffusion} ($p=4$).}
        \label{tab:abla_patch}
        \small
        \begin{tabular}{lcc}
            \toprule
            \textbf{Model} & \textbf{FID}$\downarrow$ & \textbf{IS}$\uparrow$ \\
            \midrule
            \rowcolor{myblue!60}
            \textbf{Next-Patch Diffusion}       & \textbf{1.98} & \textbf{276.7} \\
            Block-wise$\rightarrow$Full & 2.07 & 271.8 \\
            Patch$\rightarrow$Token Raster & 2.15 & 270.0 \\
            \bottomrule
        \end{tabular}
    \end{minipage}
\end{table}

\begin{figure}[t]
\vspace{1mm}
\centering
\begin{minipage}{0.62\linewidth}{
\includegraphics[width=0.95\textwidth]{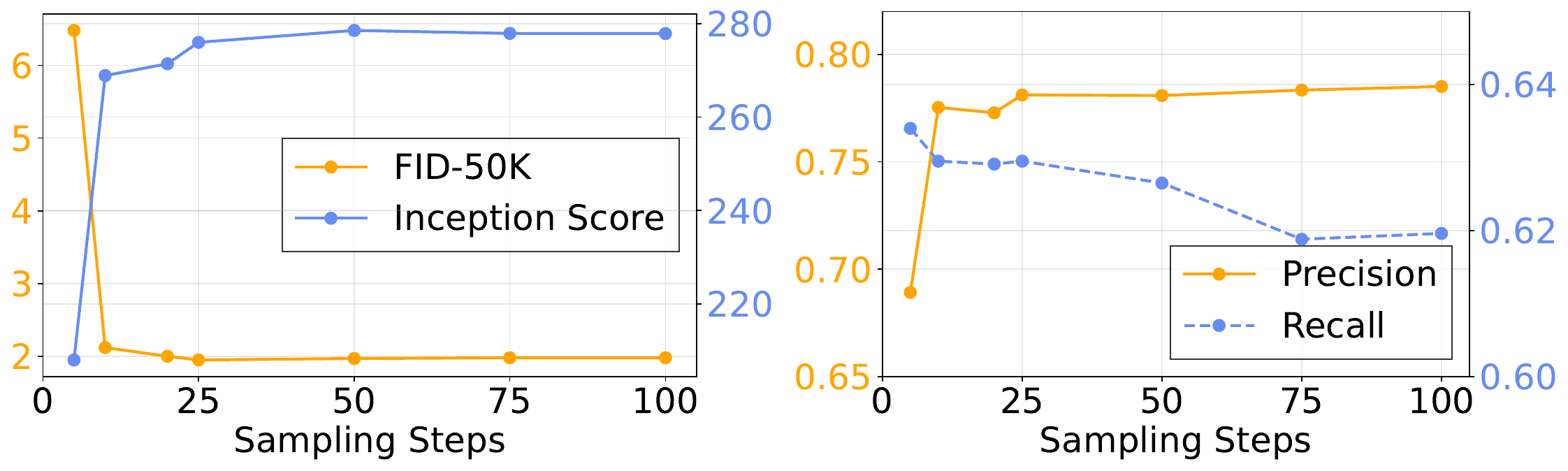}
}\end{minipage}
\hspace{-1.0em}
\begin{minipage}{0.38\linewidth}{
\caption{
\textbf{Sampling steps in binary diffusion head} ($p=4$). High-quality generation can be achieved with a small number of sampling steps. (e.g., 10-20 steps).
}
\label{fig:abla_diff_step}
}\end{minipage}
\end{figure}

\subsection{Ablation Study}
We conduct ablation studies on ImageNet 256$\times$256 to analyze the components within BitDance. We use BitDance-B as the AR generation backbone. All models are trained for 400 epochs and evaluated using their best classifier-free guidance~\cite{ho2022classifier} scale.

\textbf{Continuous VAE vs. Binary Tokenizer.} Comparing continuous VAEs~\cite{li2024autoregressive,yao2025reconstruction} against our binary tokenizer for AR generation, Tab.~\ref{tab:abla_vae} reveals notably inferior performance for the former. This suggests that unconstrained continuous tokens lead to significant error accumulation during generation.

\textbf{Sampling Head.} We evaluate different sampling heads discussed in Sec.~\ref{sec:head} in Tab.~\ref{tab:abla_head}. The token classification head suffers from OOM issues caused by a high volume of parameters, while the bitwise classification head performs poorly due to the restrictive assumption of bit independence.

\textbf{Next-Patch Diffusion.} Two critical designs in next-patch diffusion modeling are investigated: the patch-wise raster scan order for token generation and the use of block causal masks for intra-patch token visibility. As shown in Tab.~\ref{tab:abla_patch}, both designs effectively improve generation performance.

\textbf{Diffusion Sampling Steps.} Fig.~\ref{fig:abla_diff_step} illustrates how the number of sampling steps in the binary diffusion head affects performance. Notably, good results are attained with as few as 10 steps. This suggests that the discrete nature of binary tokens simplifies the sampling task relative to continuous tokens, enabling rapid and robust generation.

\begin{figure}[t!] 
\centering
    \includegraphics[width=1.0\textwidth]{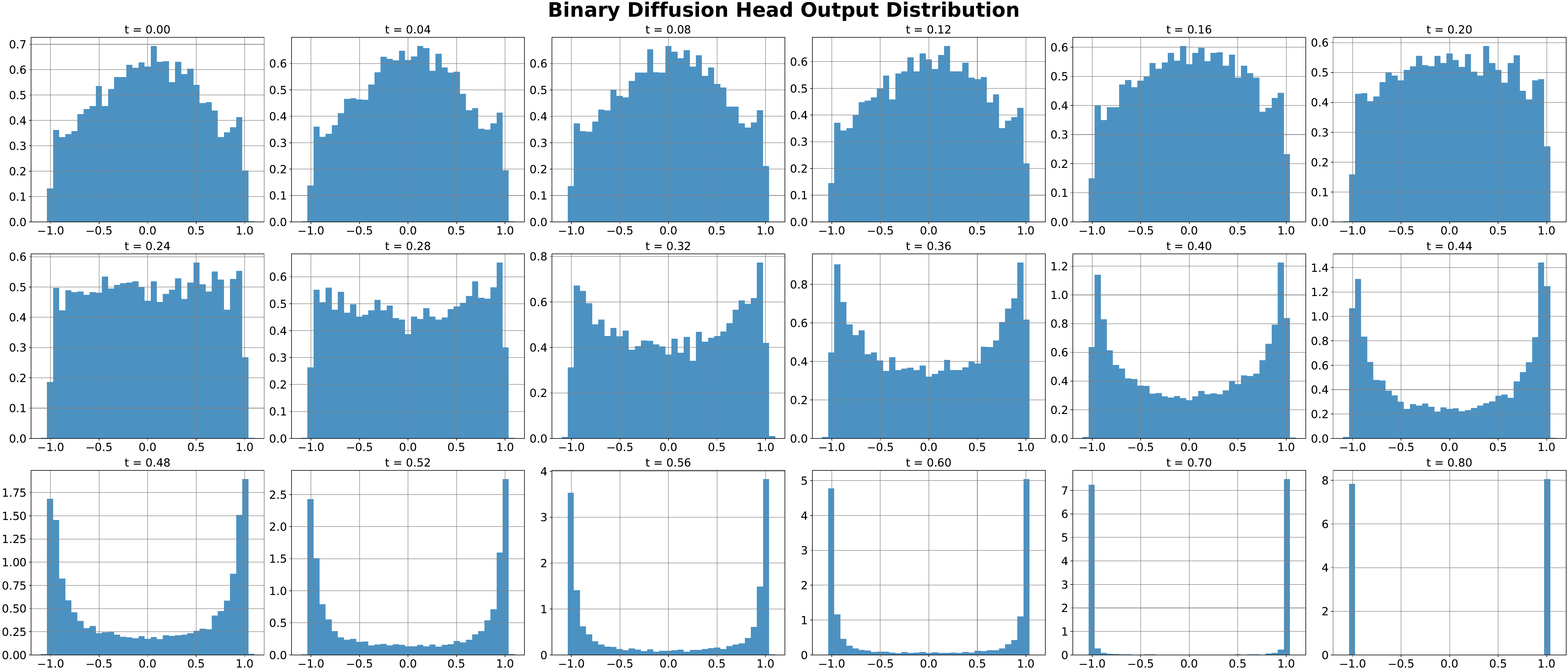}
    \caption{\textbf{Output distribution of binary diffusion head at different timesteps $t$.} As the timestep $t$ increases from 0 to 1 (i.e., as noise decreases), the predictions of the binary diffusion head become progressively more distinct, converging towards the binary values of -1 and 1.} 
    \label{fig:head_distribution}
\end{figure}

\textbf{Prediction of Binary Diffusion Head.} As described in Sec.~\ref{sec:head}, our binary diffusion head adopts the $x$-prediction~\cite{li2025back} formulation, meaning the network directly predicts the clean data. In our specific case, the prediction target is the binary latents. Fig.~\ref{fig:head_distribution} illustrates the output distribution of the head across different timesteps. When $t$ is small (indicating high noise levels), the prediction task is challenging, causing many predicted values to cluster around 0. As $t$ increases, the network's predictions become progressively more distinct, converging towards the binary values of -1 and 1. This demonstrates that our binary diffusion head implicitly learns the characteristics of a binary discrete distribution, even without explicit, hand-crafted constraints in network design or training objective.

\section{Conclusion}
We introduce BitDance, a simple yet scalable autoregressive model that achieves high-quality image generation through the efficient prediction of binary visual tokens. BitDance marks the first time a visual tokenizer’s vocabulary has been expanded to $2^{256}$, attaining reconstruction fidelity comparable to that of continuous VAEs. By leveraging the proposed binary diffusion head and next-patch diffusion modeling, BitDance effectively captures the joint distribution of multiple binary tokens, enabling efficient and precise parallel prediction. Extensive evaluations across both class-conditional and text-to-image generation benchmarks validate the effectiveness and efficiency of BitDance. We aim to further scale up the data and model size, exploring BitDance's potential in a wider range of multi-modal tasks.

\clearpage

\bibliographystyle{plainnat}
\bibliography{main}



\end{document}